\RequirePackage[loading]{tracefnt}

\documentclass[10pt,journal,compsoc]{IEEEtran}

\usepackage{multirow}

\ifCLASSOPTIONcompsoc
  \usepackage[nocompress]{cite}
\else
  \usepackage{cite}
\fi
\ifCLASSINFOpdf
  \usepackage[pdftex]{graphicx}
\usepackage{subfig}
\else
\fi
\usepackage[cmex10]{amsmath}
\usepackage{amssymb}

\usepackage{algorithmic}
\usepackage{algorithm}

\usepackage{enumitem}
\newcommand{\N}{\mathcal{N}}
\newcommand{\expectation}{\mathbb{E}}
\newcommand{\bs}{\boldsymbol}

\newcommand{\bw}{\textbf{w}}
\newcommand{\by}{\textbf{y}}
\newcommand{\bz}{\textbf{z}}

\newcommand{\specialcell}[2][c]{%
\begin{tabular}[#1]{@{}c@{}}#2\end{tabular}}

\begin{document}
%
\title{Learning Supervised Topic Models for Classification and Regression from Crowds}
%
%
%
%

\author{Filipe~Rodrigues,
        Mariana~Louren\c{c}o,
        Bernardete~Ribeiro,~\IEEEmembership{Senior~Member,~IEEE},
        and~Francisco~C.~Pereira,~\IEEEmembership{Member,~IEEE}
\IEEEcompsocitemizethanks{\IEEEcompsocthanksitem F.~Rodrigues is with the Technical University of Denmark (DTU), Bygning 115, 2800 Kgs. Lyngby, Denmark. E-mail: rodr@dtu.dk\protect
\IEEEcompsocthanksitem M.~Louren\c{c}o and B.~Ribeiro are with the CISUC, Department of Informatics Engineering, University of Coimbra, Coimbra, Portugal.\protect 
\IEEEcompsocthanksitem F.~C.~Pereira is with the Massachusetts Institute of Technology (MIT),
77 Massachusetts Avenue, 02139 Cambridge, MA, USA and Technical University of Denmark (DTU), Bygning 115, 2800 Kgs. Lyngby, Denmark. 
}
\thanks{DOI: 10.1109/TPAMI.2017.2648786}\thanks{URL: \mbox{https://ieeexplore.ieee.org/abstract/document/7807338/}}}

%
%

\markboth{IEEE Transactions on Pattern Analysis and Machine Intelligence,~Vol.~X, No.~X, XXXX}%
{Learning Supervised Topic Models for Classification and Regression from Crowds}
%



\IEEEtitleabstractindextext{%
\begin{abstract}
The growing need to analyze large collections of documents has led to great developments in topic modeling. Since documents are frequently associated with other related variables, such as labels or ratings, much interest has been placed on supervised topic models. However, the nature of most annotation tasks, prone to ambiguity and noise, often with high volumes of documents, deem learning under a single-annotator assumption unrealistic or unpractical for most real-world applications. In this article, we propose two supervised topic models, one for classification and another for regression problems, which account for the heterogeneity and biases among different annotators that are encountered in practice when learning from crowds. We develop an efficient stochastic variational inference algorithm that is able to scale to very large datasets, and we empirically demonstrate the advantages of the proposed model over state-of-the-art approaches.
\end{abstract}

}

\maketitle

\IEEEdisplaynontitleabstractindextext

%
\IEEEpeerreviewmaketitle

\ifCLASSOPTIONcompsoc
\IEEEraisesectionheading{\section{Introduction}\label{sec:introduction}}
\else
\section{Introduction}
\label{sec:introduction}
\fi

%
%
%
%

\IEEEPARstart{T}{opic} models, such as latent Dirichlet allocation (LDA), allow us to analyze large collections of documents by revealing their underlying themes, or topics, and how each document exhibits them \cite{Blei2003}. Therefore, it is not surprising that topic models have become a standard tool in data analysis, with many applications that go even beyond their original purpose of modeling textual data, such as analyzing images \cite{FeiFei2005,Wang2009}, videos \cite{Niebles2008}, survey data \cite{Erosheva2007} or social networks data \cite{Airoldi2007}. 

Since documents are frequently associated with other variables such as labels, tags or ratings, much interest has been placed on supervised topic models \cite{Mcauliffe2008}, which allow the use of that extra information to ``guide" the topics discovery. By jointly learning the topics distributions and a classification or regression model, supervised topic models have been shown to outperform the separate use of their unsupervised analogues together with an external regression/classification algorithm \cite{Wang2009,Zhu2012}. 

Supervised topics models are then state-of-the-art approaches for predicting target variables associated with complex high-dimensional data, such as documents or images. Unfortunately, the size of modern datasets makes the use of a single annotator unrealistic and unpractical for the majority of the real-world applications that involve some form of human labeling. For instance, the popular Reuters-21578 benchmark corpus was categorized by a group of personnel from Reuters Ltd and Carnegie Group, Inc. Similarly, the LabelMe project asks volunteers to annotate images from a large collection using an online tool. Hence, it is seldom the case where a single oracle labels an entire collection. 

Furthermore, the Web, through its social nature, also exploits the wisdom of crowds to annotate large collections of documents and images. By categorizing texts, tagging images or rating products and places, Web users are generating large volumes of labeled content. However, when learning supervised models from crowds, the quality of labels can vary significantly due to task subjectivity and differences in annotator reliability (or bias) \cite{Snow2008,Rodrigues2013a}. If we consider a sentiment analysis task, it becomes clear that the subjectiveness of the exercise is prone to generate considerably distinct labels from different annotators. Similarly, online product reviews are known to vary considerably depending on the personal biases and volatility of the reviewer's opinions. It is therefore essential to account for these issues when learning from this increasingly common type of data. Hence, the interest of researchers on building models that take the reliabilities of different annotators into consideration and mitigate the effect of their biases has spiked during the last few years (e.g. \cite{Welinder2010,Yan2014}). 

The increasing popularity of crowdsourcing platforms like Amazon Mechanical Turk (AMT) has further contributed to the recent advances in learning from crowds. This kind of platforms offers a fast, scalable and inexpensive solution for labeling large amounts of data. However, their heterogeneous nature in terms of contributors makes their straightforward application prone to many sorts of labeling noise and bias. Hence, a careless use of crowdsourced data as training data risks generating flawed models. 

In this article, we propose a fully generative supervised topic model that is able to account for the different reliabilities of multiple annotators and correct their biases. The proposed model is then capable of jointly modeling the words in documents as arising from a mixture of topics, the latent true target variables as a result of the empirical distribution over topics of the documents, and the labels of the multiple annotators as noisy versions of that latent ground truth. We propose two different models, one for classification \cite{Rodrigues2015} and another for regression problems, thus covering a very wide range of possible practical applications, as we empirically demonstrate. Since the majority of the tasks for which multiple annotators are used generally involve complex data such as text, images and video, by developing a multi-annotator supervised topic model we are contributing with a powerful tool for learning predictive models of complex high-dimensional data from crowds. 

Given that the increasing sizes of modern datasets can pose a problem for obtaining human labels as well as for Bayesian inference, we propose an efficient stochastic variational inference algorithm \cite{Hoffman2013} that is able to scale to very large datasets. We empirically show, using both simulated and real multiple-annotator labels obtained from AMT for popular text and image collections, that the proposed models are able to outperform other state-of-the-art approaches in both classification and regression tasks. We further show the computational and predictive advantages of the stochastic variational inference algorithm over its batch counterpart.

\section{State of the art}
\label{sec:soa}

\subsection{Supervised topic models}

Latent Dirichlet allocation (LDA) soon proved to be a powerful tool for modeling documents \cite{Blei2003} and images \cite{FeiFei2005} by extracting their underlying topics, where topics are probability distributions across words, and each document is characterized by a probability distribution across topics. However, the need to model the relationship between documents and labels quickly gave rise to many supervised variants of LDA. One of the first notable works was that of supervised LDA (sLDA) \cite{Mcauliffe2008}. By extending LDA through the inclusion of a response variable that is linearly dependent on the mean topic-assignments of the words in a document, sLDA is able to jointly model the documents and their responses, in order to find latent topics that will best predict the response variables for future unlabeled documents. Although initially developed for general continuous response variables, sLDA was later extended to classification problems \cite{Wang2009}, by modeling the relationship between topic-assignments and labels with a softmax function as in logistic regression. 

From a classification perspective, there are several ways in which document classes can be included in LDA. The most natural one in this setting is probably the sLDA approach, since the classes are directly dependent on the empirical topic mixture distributions. This approach is coherent with the generative perspective of LDA but, nevertheless, several discriminative alternatives also exist. For example, DiscLDA \cite{Lacoste2009} introduces a class-dependent linear transformation on the topic mixture proportions of each document, such that the per-word topic assignments are drawn from linearly transformed mixture proportions. The class-specific transformation matrices are then able to reposition the topic mixture proportions so that documents with the same class labels have similar topics mixture proportions. The transformation matrices can be estimated by maximizing the conditional likelihood of response variables as the authors propose \cite{Lacoste2009}. 

An alternative way of including classes in LDA for supervision is the one proposed in the Labeled-LDA model \cite{Ramage2009}. Labeled-LDA is a variant of LDA that incorporates supervision by constraining the topic model to assign to a document only topics that correspond to its label set. While this allows for multiple labels per document, it is restrictive in the sense that the number of topics needs to be the same as the number of possible labels.

From a regression perspective, other than sLDA, the most relevant approaches are the Dirichlet-multimonial regression \cite{Mimno2008} and the inverse regression topic models \cite{Rabinovich2014}. The Dirichlet-multimonial regression (DMR) topic model \cite{Mimno2008} includes a log-linear prior on the document's mixture proportions that is a function of a set of arbitrary features, such as author, date, publication venue or references in scientific articles. The inferred Dirichlet-multinomial distribution can then be used to make predictions about the values of theses features. The inverse regression topic model (IRTM) \cite{Rabinovich2014} is a mixed-membership extension of the multinomial inverse regression (MNIR) model proposed in \cite{Taddy2013} that exploits the topical structure of text corpora to improve its predictions and facilitate exploratory data analysis. However, this results in a rather complex and inefficient inference procedure. Furthermore, making predictions in the IRTM is not trivial. For example, MAP estimates of targets will be in a different scale than the original document's metadata. Hence, the authors propose the use of a linear model to regress metadata values onto their MAP predictions. 

The approaches discussed so far rely on likelihood-based estimation procedures. The work in \cite{Zhu2012} contrasts with these approaches by proposing MedLDA, a supervised topic model that utilizes the max-margin principle for estimation. Despite its margin-based advantages, MedLDA looses the probabilistic interpretation of the document classes given the topic mixture distributions. On the contrary, in this article we propose a fully generative probabilistic model of the answers of multiple annotators and of the words of documents arising from a mixture of topics. 

\subsection{Learning from multiple annotators}

Learning from multiple annotators is an increasingly important research topic. Since the early work of Dawid and Skeene \cite{DawidSkeene1979}, who attempted to obtain point estimates of the error rates of patients given repeated but conflicting responses to various medical questions, many approaches have been proposed. These usually rely on latent variable models. For example, in \cite{Smyth1995} the authors propose a model to estimate the ground truth from the labels of multiple experts, which is then used to train a classifier. 

While earlier works usually focused on estimating the ground truth and the error rates of different annotators, recent works are more focused on the problem of learning classifiers using multiple-annotator data. This idea was explored by Raykar et \mbox{al.} \cite{Raykar2010}, who proposed an approach for jointly learning the levels of expertise of different annotators and the parameters of a logistic regression classifier, by modeling the ground truth labels as latent variables. This work was later extended in \cite{Yan2014} by considering the dependencies of the annotators' labels on the instances they are labeling, and also in \cite{Rodrigues2014} through the use of Gaussian process classifiers. The model proposed in this article for classification problems shares the same intuition with this line of work and models the true labels as latent variables. However, it differs significantly by using a fully Bayesian approach for estimating the reliabilities and biases of the different annotators. Furthermore, it considers the problems of learning a low-dimensional representation of the input data (through topic modeling) and modeling the answers of multiple annotators jointly, providing an efficient stochastic variational inference algorithm. 

Despite the considerable amount of approaches for learning classifiers from the noisy answers of multiple annotators, for continuous response variables this problem has been approached in a much smaller extent. For example, Groot et \mbox{al.} \cite{Groot2011} address this problem in the context of Gaussian processes. In their work, the authors assign a different variance to the likelihood of the data points provided by the different annotators, thereby allowing them to have different noise levels, which can be estimated by maximizing the marginal likelihood of the data. Similarly, the authors in \cite{Raykar2010} propose an extension of their own classification approach to regression problems by assigning different variances to the Gaussian noise models of the different annotators. In this article, we take this idea one step further by also considering a per-annotator bias parameter, which gives the proposed model the ability to overcome certain personal tendencies in the annotators labeling styles that are quite common, for example, in product ratings and document reviews. Furthermore, we empirically validate the proposed model using real multi-annotator data obtained from Amazon Mechanical Turk. This contrasts with the previously mentioned works, which rely only on simulated annotators. 

\section{Classification model}
\label{sec:classification}

In this section, we develop a multi-annotator supervised topic model for classification problems. The model for regression settings will be presented in Section~\ref{sec:regression}. We start by deriving a (\textit{batch}) variational inference algorithm for approximating the posterior distribution over the latent variables and an algorithm to estimate the model parameters. We then develop a stochastic variational inference algorithm that gives the model the capability of handling large collections of documents. Finally, we show how to use the learned model to classify new documents.

\subsection{Proposed model}

Let $\mathcal{D} = \{\bw^d, \by^d\}_{d=1}^D$ be an annotated corpus of size $D$, where each document $\bw^d$ is given a set of labels $\by^d = \{y_r^d\}_{r=1}^{R_d}$ from $R_d$ distinct annotators. We can take advantage of the inherent topical structure of documents and model their words as arising from a mixture of topics, each being defined as a distribution over the words in a vocabulary, as in LDA. In LDA, the $n^{th}$ word, $w_n^d$, in a document $d$ is provided a discrete topic-assignment $z_n^d$, which is drawn from the documents' distribution over topics $\theta^d$. This allows us to build lower-dimensional representations of documents, which we can explore to build classification models by assigning coefficients $\bs\eta$ to the mean topic-assignment of the words in the document, $\bar{z}^d$, and applying a softmax function in order to obtain a distribution over classes. Alternatively, one could consider more flexible models such as Gaussian processes, however that would considerably increase the complexity of inference. 

Unfortunately, a direct mapping between document classes and the labels provided by the different annotators in a multiple-annotator setting would correspond to assuming that they are all equally reliable, an assumption that is violated in practice, as previous works clearly demonstrate (e.g. \cite{Snow2008,Rodrigues2013a}). Hence, we assume the existence of a latent ground truth class, and model the labels from the different annotators using a noise model that states that, given a true class $c$, each annotator $r$ provides the label $l$ with some probability $\pi_{c,l}^r$. Hence, by modeling the matrix $\bs\pi^r$ we are in fact modeling a per-annotator (normalized) confusion matrix, which allows us to account for their different levels of expertise and correct their potential biases. 

The generative process of the proposed model for classification problems can then be summarized as follows:

\begin{enumerate}[leftmargin=0.7cm]
\item For each annotator $r$
\begin{enumerate}
\item For each class $c$
\begin{enumerate}
\item Draw reliability parameter  $\pi_c^r | \omega \sim \mbox{Dir}(\omega)$
\end{enumerate}
\end{enumerate}
\item For each topic $k$
\begin{enumerate}
\item Draw topic distribution $\beta_k | \tau \sim \mbox{Dir}(\tau)$
\end{enumerate}
\item For each document $d$
\begin{enumerate}
\item Draw topic proportions $\theta^d | \alpha \sim \mbox{Dir}(\alpha)$
\item For the $n^{th}$ word
\begin{enumerate}
\item Draw topic assignment $z_n^d | \theta^d \sim \mbox{Mult}(\theta^d)$
\item Draw word $w_n^d | z_n^d, \bs{\beta} \sim \mbox{Mult}(\beta_{z_n^d})$
\end{enumerate}
\item Draw latent (true) class $c^d | \bz^d, \bs\eta \sim \mbox{Softmax}(\bar{z}^d,\bs\eta)$
\item For each annotator $r \in R_d$
\begin{enumerate}
\item Draw annotator's label $y^{d,r}| c^d, \bs{\pi}^r \sim \mbox{Mult}(\pi_{c^d}^r)$
\end{enumerate}
\end{enumerate}
\end{enumerate}
where $R_d$ denotes the set of annotators that labeled the $d^{th}$ document, $\bar{z}^d = \frac{1}{N_d} \sum_{n=1}^{N_d} z_n^d$, and the softmax is given by
\begin{align}
p(c^d|\bar{z}^d,\bs\eta) = \frac{\exp(\eta_c^T \bar{z}^d)}{\sum_{l=1}^C \exp(\eta_l^T \bar{z}^d)}.\nonumber
\end{align}

\begin{figure}[t]
\centering
\includegraphics[width=8.4cm]{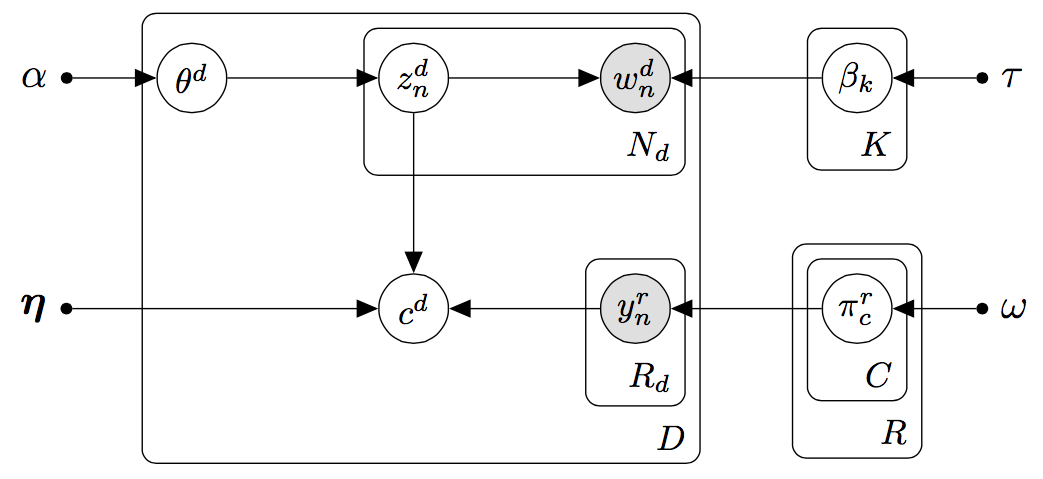}
\caption{Graphical representation of the proposed model for classification.}
\label{fig:graphical_model}
\end{figure}

Fig.~\ref{fig:graphical_model} shows a graphical model representation of the proposed model, where $K$ denotes the number of topics, $C$ is the number of classes, $R$ is the total number of annotators and $N_d$ is the number of words in the document $d$. Shaded nodes are used to distinguish latent variable from the observed ones and small solid circles are used to denote model parameters. Notice that we included a Dirichlet prior over the topics $\beta_k$ to produce a smooth posterior and control sparsity. Similarly, instead of computing maximum likelihood or MAP estimates for the annotators reliability parameters $\pi_c^r$, we place a Dirichlet prior over these variables and perform approximate Bayesian inference. This contrasts with previous works on learning classification models from crowds \cite{Raykar2010,Yan2010}. 

\subsection{Approximate inference}

\label{subsec:approx_inf}

Given a dataset $\mathcal{D}$, the goal of inference is to compute the posterior distribution of the per-document topic proportions $\theta^d$, the per-word topic assignments $z_n^d$, the per-topic distribution over words $\beta_k$, the per-document latent true class $c^d$, and the per-annotator confusion parameters $\bs\pi^r$. As with LDA, computing the exact posterior distribution of the latent variables is computationally intractable. Hence, we employ mean-field variational inference to perform approximate Bayesian inference. 

Variational inference methods seek to minimize the KL divergence between the variational and the true posterior distribution. We assume a fully-factorized (mean-field) variational distribution of the form
\begin{align}
q(\bs{\theta}, \bz_{1:D}, \bs{c}, \bs{\beta}, \boldsymbol{\pi}_{1:R}) &= \bigg(\prod_{r=1}^R \prod_{c=1}^C q(\pi_c^r|\xi_c^r)\bigg) \bigg(\prod_{i=1}^K q (\beta_i | \zeta_i)\bigg) \nonumber\\
&\times \prod_{d=1}^D q(\theta^d|\gamma^d) \, q(c^d|\lambda^d) \prod_{n=1}^{N_d} q(z_n^d | \phi_n^d),\nonumber
\end{align}
where $\boldsymbol{\xi}_{1:R}$, $\bs{\zeta}$, $\bs{\gamma}$, $\bs{\lambda}$ and $\boldsymbol{\phi}_{1:D}$ are variational parameters. Table \ref{table:variational_parameters} shows the correspondence between variational parameters and the original parameters.


\begin{table}[t]
\caption{Correspondence between variational parameters and the original parameters.}
\label{table:variational_parameters}
\begin{center}
\begin{tabular}{c|c|c|c|c|c}
Original parameter & $\pi_c^r$ & $\beta_k$ & $\theta^d$ & $c^d$ & $z_n^d$\\
Variational parameter & $\xi_c^r$ & $\zeta_k$ & $\gamma^d$ & $\lambda^d$ & $\phi_n^d$ 
\end{tabular}
\end{center}
\vspace*{-0.7cm}
\end{table}%

Let $\Theta = \{\alpha,\tau,\omega,\bs\eta\}$ denote the model parameters. Following \cite{Jordan1999}, the KL minimization can be equivalently formulated as maximizing the following lower bound on the log marginal likelihood
\begin{align}
\log\,p(&\textbf{w}_{1:D},\textbf{y}_{1:D}|\Theta)  \nonumber\\
&=\log \int \sum_{\bz,\bs{c}} q(\bs{\theta}, \textbf{z}_{1:D}, \bs{c}, \bs{\beta}, \boldsymbol{\pi}_{1:R})\nonumber\\
&\times \frac{p(\bs{\theta}, \textbf{z}_{1:D}, \bs{c}, \textbf{w}_{1:D}, \textbf{y}_{1:D}, \bs{\beta},  \boldsymbol{\pi}_{1:R}|\Theta)}{q(\bs{\theta}, \textbf{z}_{1:D}, \bs{c}, \bs{\beta}, \boldsymbol{\pi}_{1:R})} \, d\bs\theta \, d\bs\beta \, d\bs\pi_{1:R}\nonumber\\
&\geqslant \expectation_q[\log p(\bs{\theta}, \textbf{z}_{1:D}, \bs{c}, \textbf{w}_{1:D}, \textbf{y}_{1:D}, \bs{\beta},  \boldsymbol{\pi}_{1:R}|\Theta)]\nonumber\\
&- \expectation_q[\log q(\bs{\theta}, \textbf{z}_{1:D}, \bs{c}, \bs{\beta}, \boldsymbol{\pi}_{1:R})]\nonumber\\
&= \mathcal{L}(\bs{\gamma}, \boldsymbol{\phi}_{1:D}, \bs{\lambda}, \bs{\zeta}, \boldsymbol{\xi}_{1:R}|\Theta),
\label{eq:lowerbound}
\end{align}
which we maximize using coordinate ascent. 

Optimizing $\mathcal{L}$ \mbox{w.r.t.} $\gamma$ and $\zeta$ gives the same coordinate ascent updates as in LDA \cite{Blei2003}
\begin{align}
\label{eq:gamma_update}
\gamma_i^d &= \alpha + \sum_{n=1}^{N_d} \phi_{n,i}^d ,\\
\label{eq:zeta_update}
\zeta_{i,j} &= \tau + \sum_{d=1}^D\sum_{n=1}^{N_d} w_{n,j}^d  \phi_{n,i}^d.
\end{align}

The variational Dirichlet parameters $\xi$ can be optimized by collecting only the terms in $\mathcal{L}$ that contain $\xi$
\begin{align}
\mathcal{L}_{[\xi]} &= \sum_{r=1}^R\sum_{c=1}^C\sum_{l=1}^C \expectation_q[\log \pi_{c,l}^r]  \bigg( \omega + \sum_{d=1}^{D_r}\lambda_c^d y_l^{d,r} - \xi_{c,l}^r \bigg) \nonumber\\
&-\sum_{r=1}^R\sum_{c=1}^C \log \Gamma \bigg( \sum_{t=1}^C \xi_{c,t}^r\bigg) +\sum_{r=1}^R\sum_{c=1}^C\sum_{l=1}^C \log \Gamma(\xi_{c,l}^r),\nonumber
\end{align}
where $D_r$ denotes the documents labeled by the $r^{th}$ annotator, $\expectation_q[\log \pi_{c,l}^r] = \Psi(\xi_{c,l}^r) - \Psi(\sum_{t=1}^C \xi_{c,t}^r)$, and $\Gamma(\cdot)$ and $\Psi(\cdot)$ are the gamma and digamma functions, respectively. Taking derivatives of $\mathcal{L}_{[\xi]}$ \mbox{w.r.t.} $\xi$ and setting them to zero, yields the following update
\begin{align}
\label{eq:xi_update}
\xi_{c,l}^r &= \omega + \sum_{d=1}^{D_r}\lambda_c^d y_l^{d,r}.
\end{align}

Similarly, the coordinate ascent updates for the documents distribution over classes $\lambda$ can be found by considering the terms in $\mathcal{L}$ that contain $\lambda$
\begin{align}
\mathcal{L}_{[\lambda]} &= \sum_{d=1}^D\sum_{l=1}^C \lambda_l^d \eta_l^T \bar{\phi}^d - \sum_{l=1}^C \lambda_l^d \log \lambda_l^d\nonumber\\
&+ \sum_{d=1}^D \sum_{r=1}^{R_d} \sum_{l=1}^C \sum_{c=1}^C\lambda_l^d \, y_c^{d,r} \, \expectation_q[\log \pi_{l,c}^r],\nonumber
\end{align}
where $\bar{\phi}^d = \frac{1}{N_d} \sum_{n=1}^{N_d} \phi_n^d$.
Adding the necessary Lagrange multipliers to ensure that $\sum_{l=1}^C \lambda_l^d = 1$ and setting the derivatives \mbox{w.r.t.} $\lambda_l^d$ to zero gives the following update
\begin{align}
\lambda_l^d &\propto \exp\bigg( \eta_l^T \bar{\phi}^d + \sum_{r=1}^{R_d} \sum_{c=1}^C y_c^{d,r} \, \expectation_q[\log \pi_{l,c}^r]\bigg).
\label{eq:lambda_update}
\end{align}
Observe how the variational distribution over the true classes results from a combination between the dot product of the inferred mean topic assignment $\bar{\phi}^d$ with the coefficients $\bs\eta$ and the labels $\by$ from the multiple annotators ``weighted" by their expected log probability $\expectation_q[\log \pi_{l,c}^r]$.

The main difficulty of applying standard variational inference methods to the proposed model is the non-conjugacy between the distribution of the mean topic-assignment $\bar{z}^d$ and the softmax. Namely, in the expectation
\begin{align}
\expectation_q[\log p(c^d|\bar{z}^d,\bs\eta)] = \expectation_q[\eta_{c^d}^T \bar{z}^d] - \expectation_q\Big[\log \sum_{l=1}^C \exp(\eta_l^T \bar{z}^d)\Big],\nonumber
\end{align}
the second term is intractable to compute. We can make progress by applying Jensen's inequality to bound it as follows
\begin{align}
-  \expectation_q\bigg[\log \sum_{l=1}^C \exp(\eta_l^T \bar{z}^d)\bigg] &\geqslant - \log  \sum_{l=1}^C \expectation_q[\exp(\eta_l^T \bar{z}^d)] \nonumber\\
&= - \log  \sum_{l=1}^C \prod_{j=1}^{N_d} \big( \phi_j^d \big)^T \exp\Big(\eta_l \frac{1}{N_d}\Big)\nonumber\\
&= - \log ( a^T \phi_n^d ),\nonumber
\end{align}
where $a \triangleq \sum_{l=1}^C \exp (\frac{\eta_l}{N_d}) \prod_{j=1,j \neq n}^{N_d} \big( \phi_j^d \big)^T \exp\Big(\frac{\eta_l}{N_d}\Big)$, which is constant \mbox{w.r.t.} $\phi_n^d$. This local variational bound can be made tight by noticing that $\log(x) \leqslant \epsilon^{-1} x + \log(\epsilon) - 1, \forall x > 0, \epsilon > 0$, where equality holds if and only if $x = \epsilon$. Hence, given the current parameter estimates $(\phi_n^d)^{old}$, if we set $x = a^T \phi_n^d$ and $\epsilon = a^T (\phi_n^d)^{old}$ then, for an individual parameter $\phi_n^d$, we have that
\begin{align}
-  \expectation_q\bigg[\log &\sum_{l=1}^C \exp(\eta_l^T \bar{z}^d)\bigg] \nonumber\\
&\geqslant - (a^T (\phi_n^d)^{old})^{-1} (a^T \phi_n^d) - \log(a^T (\phi_n^d)^{old}) + 1\nonumber.
\end{align}
Using this local bound to approximate the expectation of the log-sum-exp term, and taking derivatives of the evidence lower bound \mbox{w.r.t.} $\phi_n$ with the constraint that $\sum_{i=1}^K \phi_{n,i}^d = 1$, yields the following fix-point update
\begin{align}
\phi_{n,i}^d &\propto \exp \Bigg( \Psi(\gamma_i)  + \sum_{j=1}^V w_{n,j}^d \bigg(\Psi(\zeta_{i,j}) - \Psi\bigg(\sum_{k=1}^V \zeta_{i,k}\bigg) \bigg) \nonumber\\
&+ \frac{\sum_{l=1}^C \lambda_l^d \eta_{l,i}}{N_d}  - (a^T (\phi_n^d)^{old})^{-1} a_i\Bigg).
\label{eq:phi_update}
\end{align}
where $V$ denotes the size of the vocabulary. Notice how the per-word variational distribution over topics $\phi$ depends on the variational distribution over the true class label $\lambda$. 

The variational inference algorithm iterates between Eqs.~\ref{eq:gamma_update}-\ref{eq:phi_update} until the evidence lower bound, \mbox{Eq.} \ref{eq:lowerbound}, converges. Additional details are provided as supplementary material\footnote{Supplementary material available at:\\ http://www.fprodrigues.com/maslda-supp-mat.pdf}. 

\subsection{Parameter estimation}

The model parameters are $\Theta = \{\alpha,\tau,\omega,\bs\eta\}$. The parameters $\{\alpha,\tau,\omega\}$ of the Dirichlet priors can be regarded as hyper-parameters of the proposed model. As with many works on topic models (e.g. \cite{chuang2013topic,Wang2009}), we assume hyper-parameters to be fixed, since they can be effectively selected by grid-search procedures which are able to explore well the parameter space without suffering from local optima. Our focus is then on estimating the coefficients $\bs\eta$ using a variational EM algorithm. Therefore, in the E-step we use the variational inference algorithm from section~\ref{subsec:approx_inf} to estimate the posterior distribution of the latent variables, and in the M-step we find maximum likelihood estimates of $\bs\eta$ by maximizing the evidence lower bound $\mathcal{L}$. Unfortunately, taking derivatives of $\mathcal{L}$ \mbox{w.r.t.} $\bs\eta$ does not yield a closed-form solution. Hence, we use a numerical method, namely L-BFGS \cite{Nocedal06}, to find an optimum. The objective function and gradients are given by
\begin{align}
\mathcal{L}_{[\eta]} &= \sum_{d=1}^D \Bigg( \sum_{l=1}^{C} \lambda_l^d \eta_{l}^T \bar{\phi}^d - \log  \sum_{l=1}^C b_{l}^d \Bigg) \nonumber\\
\nabla_{\eta_{l,i}} &= \sum_{d=1}^D \Bigg( \lambda_{l,i}^d \bar{\phi}_i^d - \frac{b_{l}^d }{ \sum_{t=1}^C b_{t}^d}  \sum_{n=1}^{N_d} \frac{ \frac{1}{N_d} \phi_{n,i}^d \exp( \frac{1}{N_d} \eta_{l,i} ) }{\sum_{j=1}^K \phi_{n,j}^d \exp( \frac{1}{N_d} \eta_{l,j} ) }\Bigg)\nonumber
\end{align}
where, for convenience, we defined the following variable: $b_{l}^d \triangleq \prod_{n=1}^{N_d} \Big( \sum_{i=1}^K \phi_{n,i}^d \exp\Big(\frac{1}{N_d}\eta_{l,i} \Big) \Big)$.

\subsection{Stochastic variational inference}

In Section~\ref{subsec:approx_inf}, we proposed a batch coordinate ascent algorithm for doing variational inference in the proposed model. This algorithm iterates between analyzing every document in the corpus to infer the local hidden structure, and estimating the global hidden variables. However, this can be inefficient for large datasets, since it requires a full pass through the data at each iteration before updating the global variables. In this section, we develop a stochastic variational inference algorithm \cite{Hoffman2013}, which follows noisy estimates of the gradients of the evidence lower bound $\mathcal{L}$. 

Based on the theory of stochastic optimization \cite{robbins1951}, we can find unbiased estimates of the gradients by subsampling a document (or a mini-batch of documents) from the corpus, and using it to compute the gradients as if that document was observed $D$ times. Hence, given an uniformly sampled document $d$, we use the current posterior distributions of the global latent variables, $\bs\beta$ and $\bs\pi_{1:R}$, and the current coefficient estimates $\bs\eta$, to compute the posterior distribution over the local hidden variables $\theta^d$, $\bz^d$ and $c^d$ using Eqs.~\ref{eq:gamma_update}, \ref{eq:phi_update} and \ref{eq:lambda_update} respectively. These posteriors are then used to update the global variational parameters, $\bs\zeta$ and $\bs\xi_{1:R}$ by taking a step of size $\rho_t$ in the direction of the noisy estimates of the natural gradients. 

Algorithm~\ref{alg:svi} describes a stochastic variational inference algorithm for the proposed model. Given an appropriate schedule for the learning rates $\{\rho_t\}$, such that $\sum_t \rho_t$ and $\sum_t \rho_t^2 < \infty$, the stochastic optimization algorithm is guaranteed to converge to a local maximum of the evidence lower bound \cite{robbins1951}. 

\begin{algorithm}[t]
\caption{Stochastic variational inference for the proposed classification model}\label{alg:svi}
\begin{algorithmic}[1]
\STATE Initialize $\bs\gamma^{(0)}$, $\bs\phi_{1:D}^{(0)}$, $\bs\lambda^{(0)}$, $\bs\zeta^{(0)}$, $\bs\xi^{(0)}_{1:R}$, $t=0$
\REPEAT
 \STATE Set t = t + 1
 \STATE Sample a document $\bw^d$ uniformly from the corpus
 \REPEAT
 \STATE Compute $\phi_n^d$ using \mbox{Eq.}~\ref{eq:phi_update}, for $n \in \{1..N_d\}$
 \STATE Compute $\gamma^d$ using \mbox{Eq.}~\ref{eq:gamma_update}
 \STATE Compute $\lambda^d$ using \mbox{Eq.}~\ref{eq:lambda_update}
 \UNTIL local parameters $\phi_n^d$, $\gamma^d$ and $\lambda^d$ converge
 \STATE Compute step-size $\rho_t = (t + delay)^{-\kappa}$
 \STATE Update topics variational parameters
 \begin{align}
 {\zeta_{i,j}}^{(t)} &= (1-\rho_t) \, \zeta_{i,j}^{(t-1)} + \rho_t \bigg(\tau + D \sum_{n=1}^{N_d} w_{n,j}^d  \phi_{n,i}^d\bigg)\nonumber
 \end{align}
 \STATE Update annotators confusion parameters
 \begin{align}
 {\xi_{c,l}^r}^{(t)} &= (1-\rho_t) \, {\xi_{c,l}^r}^{(t-1)} + \rho_t \big(\omega + D \, \lambda_c^d \,y_l^{d,r}\big)\nonumber
 \end{align}
\UNTIL global convergence criterion is met
\end{algorithmic}
\end{algorithm}

\subsection{Document classification}

In order to make predictions for a new (unlabeled) document $d$, we start by computing the approximate posterior distribution over the latent variables $\theta^d$ and $\bz^d$. This can be achieved by dropping the terms that involve $y$, $c$ and $\pi$ from the model's joint distribution (since, at prediction time, the multi-annotator labels are no longer observed) and averaging over the estimated topics distributions. Letting the topics distribution over words inferred during training be $q(\bs\beta|\bs\zeta)$, the joint distribution for a single document is now simply given by
\begin{align}
p(\theta^d,\bz^d) &= \int q(\bs\beta|\bs\zeta) \, p(\theta^d|\alpha) \prod_{n=1}^{N_d} p(z_n^d|\theta^d) \, p(w_n^d|z_n^d,\bs\beta) \, d\bs\beta. \nonumber
\end{align}
Deriving a mean-field variational inference algorithm for computing the posterior over $q(\theta^d,\bz^d) = q(\theta^d|\gamma^d) \prod_{n=1}^{N_d} q(z_n^d|\phi_n^d)$ results in the same fixed-point updates as in LDA \cite{Blei2003} for $\gamma_i^d$ (\mbox{Eq.}~\ref{eq:gamma_update}) and $\phi_{n,i}^d$
\begin{align}
\phi_{n,i}^d &\propto \exp \Bigg( \Psi(\gamma_i)  + \sum_{j=1}^V w_{n,j}^d \bigg(\Psi(\zeta_{i,j}) - \Psi\bigg(\sum_{k=1}^V \zeta_{i,k}\bigg) \bigg) \Bigg).
\label{eq:phi_update_lda}
\end{align}
Using the inferred posteriors and the coefficients $\bs\eta$ estimated during training, we can make predictions as follows
\begin{align}
c_*^d = \arg \max_{c} \eta_c^T \bar{\phi}^d.
\end{align}
This is equivalent to making predictions in the classification version of sLDA \cite{Wang2009}.

\section{Regression model}
\label{sec:regression}

In this section, we develop a variant of the model proposed in Section~\ref{sec:classification} for regression problems. We shall start by describing the proposed model with a special focus on the how to handle multiple annotators with different biases and reliabilities when the target variables are continuous variables. Next, we present a variational inference algorithm, highlighting the differences to the classification version. Finally, we show how to optimize the model parameters.

\subsection{Proposed model}

\begin{figure}[t]
\centering
\includegraphics[width=7.7cm]{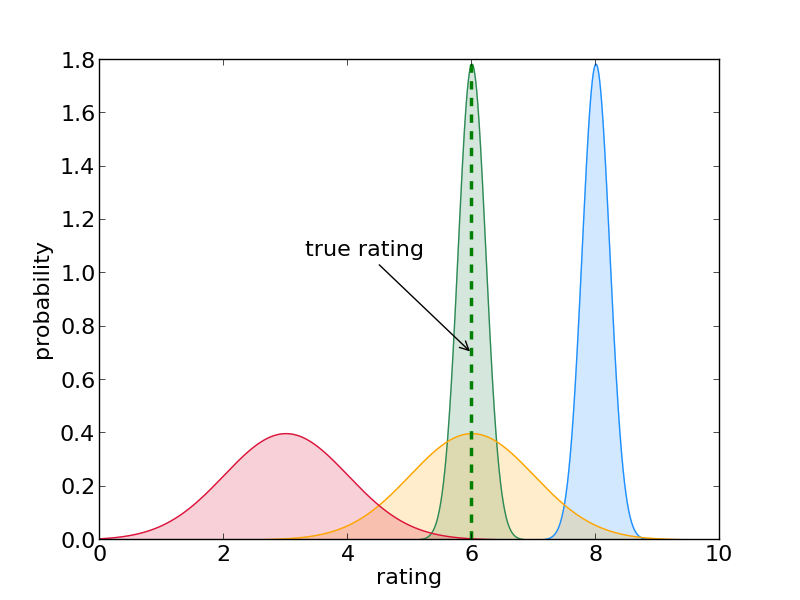}
\caption{Example of 4 different annotators (represented by different colours) with different biases and precisions.}
\label{fig:annotators_example}
\end{figure}

For developing a multi-annotator supervised topic model for regression, we shall follow a similar intuition as the one we considered for classification. Namely, we shall assume that, for a given document $d$, each annotator provides a noisy version, $y^{d,r} \in \mathbb{R}$, of the true (continuous) target variable, which we denote by $x^d \in \mathbb{R}$. This can be, for example, the true rating of a product or the true sentiment of a document. Assuming that each annotator $r$ has its own personal bias $b^r$ and precision $p^r$ (inverse variance), and assuming a Gaussian noise model for the annotators' answers, we have that
\begin{align}
y^{d,r} \sim \N(y^{d,r} | x^d + b^r, 1/p^r).
\end{align}
This approach is therefore more powerful than previous works \cite{Raykar2010,Groot2011}, where a single precision parameter was used to model the annotators' expertise. Fig.~\ref{fig:annotators_example} illustrates this intuition for 4 annotators, represented by different colors. The ``green annotator" is the best one, since he is right on the target and his answers vary very little (low bias, high precision). The ``yellow annotator" has a low bias, but his answers are very uncertain, as they can vary a lot. Contrarily, the ``blue annotator" is very precise, but consistently over-estimates the true target (high bias, high precision). Finally, the ``red annotator" corresponds to the worst kind of annotator (with high bias and low precision). 

\begin{figure}[t]
\centering
\includegraphics[width=8.3cm]{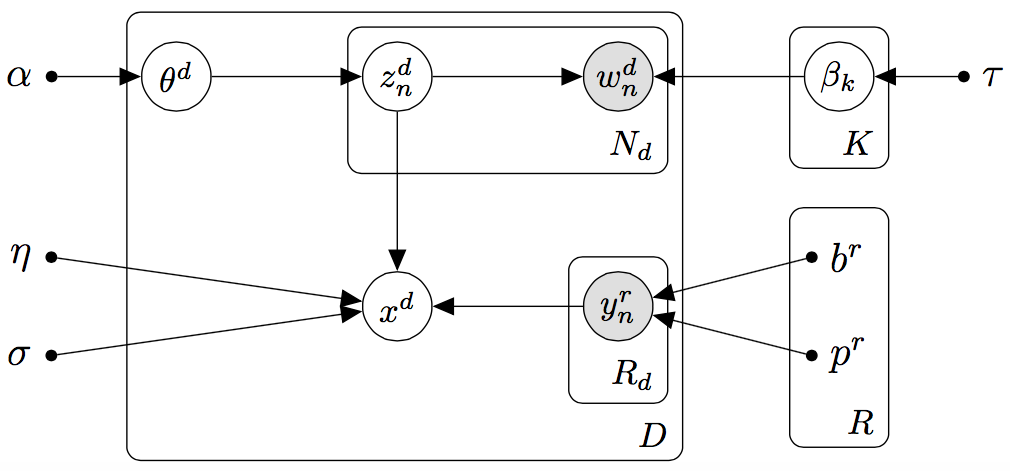}
\caption{Graphical representation of the proposed model for regression.}
\label{fig:graphical_model_reg}
\vspace{-0.4cm}
\end{figure}

Having specified a model for annotators answers given the true targets, the only thing left is to do is to specify a model of the latent true targets $x^d$ given the empirical topic mixture distributions $\bar{z}^d$. For this, we shall keep things simple and assume a linear model as in sLDA \cite{Mcauliffe2008}. The generative process of the proposed model for continuous target variables can then be summarized as follows:
\begin{enumerate}[leftmargin=0.7cm]
\item For each annotator $r$
\begin{enumerate}
\item For each class $c$
\begin{enumerate}
\item Draw reliability parameter  $\pi_c^r | \omega \sim \mbox{Dir}(\omega)$
\end{enumerate}
\end{enumerate}
\item For each topic $k$
\begin{enumerate}
\item Draw topic distribution $\beta_k | \tau \sim \mbox{Dir}(\tau)$
\end{enumerate}
\item For each document $d$
\begin{enumerate}
\item Draw topic proportions $\theta^d | \alpha \sim \mbox{Dir}(\alpha)$
\item For the $n^{th}$ word
\begin{enumerate}
\item Draw topic assignment $z_n^d | \theta^d \sim \mbox{Mult}(\theta^d)$
\item Draw word $w_n^d | z_n^d, \bs{\beta} \sim \mbox{Mult}(\beta_{z_n^d})$
\end{enumerate}
\item Draw latent (true) target $x^d | \bz^d, \eta, \sigma \sim \N(\eta^T \bar{z}^d, \sigma^2)$
\item For each annotator $r \in R_d$
\begin{enumerate}
\item Draw answer $y^{d,r}| x^d, b^r, p^r\sim \N(x^d + b^r, 1/p^r)$
\end{enumerate}
\end{enumerate}
\end{enumerate}
Fig.~\ref{fig:graphical_model_reg} shows a graphical representation of the proposed model.

\subsection{Approximate inference}
\label{subsec:approx_inf_reg}

The goal of inference is to compute the posterior distribution of the per-document topic proportions $\theta^d$, the per-word topic assignments $z_n^d$, the per-topic distribution over words $\beta_k$ and the per-document latent true targets $x^d$. As we did for the classification model, we shall develop a variational inference algorithm using coordinate ascent. The lower-bound on the log marginal likelihood is now given by
\begin{align}
\mathcal{L}(\bs{\gamma}, \boldsymbol{\phi}_{1:D}, \textbf{m}, \textbf{v}, \bs{\zeta}|\Theta) &= \expectation_q[\log p(\bs{\theta}, \textbf{z}_{1:D}, \bs{x}, \textbf{w}_{1:D}, \textbf{y}_{1:D}, \bs{\beta}|\Theta)]\nonumber\\
&- \expectation_q[\log q(\bs{\theta}, \textbf{z}_{1:D}, \bs{x}, \bs{\beta})],
\label{eq:lowerbound_reg}
\end{align}
where $\Theta = \{\alpha,\tau,\eta,\sigma,\textbf{b},\textbf{p}\}$ are the model parameters. We assume a fully-factorized (mean-field) variational distribution $q$ of the form
\begin{align}
q(\bs{\theta}, \bz_{1:D}, \bs{c}, \bs{\beta}) &= \bigg(\prod_{i=1}^K q (\beta_i | \zeta_i)\bigg) \prod_{d=1}^D q(x^d|m^d,v^d) \, q(\theta^d|\gamma^d) \nonumber\\
&\times \prod_{n=1}^{N_d} q(z_n^d | \phi_n^d).
\end{align}
where $\bs{\zeta}$, $\textbf{m}$, $\textbf{v}$, $\bs{\gamma}$ and $\boldsymbol{\phi}_{1:D}$ are the variational parameters. Notice the new Gaussian term, $q(x^d|m^d,v^d)$, corresponding to the approximate posterior distribution of the unobserved true targets. 

Optimizing the variational objective $\mathcal{L}$ \mbox{w.r.t.} $\bs\gamma$ and $\bs\zeta$ yields the same updates from Eqs.~\ref{eq:gamma_update} and \ref{eq:zeta_update}. Optimizing \mbox{w.r.t.} $\phi$ gives a similar update to the one in sLDA \cite{Mcauliffe2008}
\begin{align}
\phi_{n,i}^d &\propto \exp \Bigg( \Psi(\gamma_i) + \sum_{j=1}^V w_{n,j}^d \Bigg( \Psi(\zeta_{i,j}) - \Psi\bigg(\sum_{k=1}^V \zeta_{i,k}\bigg) \Bigg) \nonumber\\
&+\frac{m^d}{N^d\sigma^2}\eta - \frac{2 (\eta^T \phi_{-n}^d) \eta + (\eta \circ \eta)}{2 (N^d)^2 \sigma^2}\Bigg),
\label{eq:phi_update_reg}
\end{align} 
where we defined $\phi_{-n}^d \triangleq \sum_{m \neq n} \phi_m^d$. Notice how this update differs only from the one in \cite{Mcauliffe2008} by replacing the true target variable by its expected value under the variational distribution, which is given by $\expectation_q[x^d] = m^d$.

The only variables left for doing inference on are then the latent true targets $\textbf{x} = \{x^d\}_{d=1}^D$. The variational distribution of  $x^d$ is governed by two parameters: a mean $m^d$ and a variance $v^d$. Collecting all the terms in $\mathcal{L}$ that contain $m$ gives
\begin{align}
\mathcal{L}_{[m]} &= - \sum_{d=1}^D \sum_{r=1}^{R_d} \frac{p^r}{2} \bigg( (m^d)^2 + 2 m^d b^r - 2 y^{d,r} m^d  \bigg) \nonumber\\
&- \sum_{d=1}^D \frac{1}{2\sigma^2}\bigg( (m^d)^2 - 2 m^d (\eta^T \bar{\phi}^d) \bigg).
\end{align}
Taking derivatives of $\mathcal{L}_{[m]}$ and setting them to zero gives the following update for $m^d$
\begin{align}
m^d = \frac{\sigma^{-2} (\eta^T \bar{\phi}^d) + \sum_{r=1}^{R_d} p^r (y^{d,r}-b^r)}{\sigma^{-2} + \sum_{r=1}^R p^r}.
\label{eq:m_update}
\end{align}
Notice how the value of $m^d$ is a weighted average of what the linear regression model on the empirical topic mixture believes the true target should be, and the bias-corrected answers of the different annotators weighted by their individual precisions.  

As for $m$, we can optimize $\mathcal{L}$ \mbox{w.r.t.} $v$ by collecting all terms that contain $v$
\begin{align}
\mathcal{L}_{[v]} &= \sum_{d=1}^D \Bigg( \frac{1}{2} \log(v^d) - \sum_{r=1}^{R_d} \frac{p^r v^d}{2} - \frac{v^d}{2\sigma^2} \Bigg),
\end{align}
and taking derivatives, yielding the update
\begin{align}
v^d &= \sigma^2 + \sum_{r=1}^{R_d} \frac{1}{p^r}.
\label{eq:v_update}
\end{align}

\subsection{Parameter estimation}

The parameters of the proposed regression model are $\Theta = \{\alpha,\tau,\eta,\sigma,\textbf{b},\textbf{p}\}$. As we did for the classification model, we shall assume the Dirichlet parameters, $\alpha$ and $\tau$, to be fixed. Similarly, we shall assume that the variance of the true targets, $\sigma^2$, to be constant. The only parameters left to estimate are then the regression coefficients $\eta$ and the annotators biases, $\textbf{b} = \{b^r\}_{r=1}^R$, and precisions, $\textbf{p} = \{p^r\}_{r=1}^R$, which we estimate using variational Bayesian EM.

Since the latent true targets are now linear functions of the documents' empirical topic mixtures (i.e. there is no softmax function), we can find a closed form solution for the regression coefficients $\eta$. Taking derivatives of $\mathcal{L}$ \mbox{w.r.t.} $\eta$ and setting them to zero, gives the following solution for $\eta$
\begin{align}
\eta^T &= \sum_{d=1}^D \expectation_q\big[ \bar{z}^d(\bar{z}^d)^T\big]^{-1} (\bar{\phi}^d)^T m^d,
\end{align}
where
\begin{align}
\expectation_q\big[ \bar{z}^d(\bar{z}^d)^T\big] &= \frac{1}{(N^d)^2}\Bigg(\sum_{n=1}^{N^d}\sum_{m\not=n}^{N^d} \phi^d_n (\phi_m^d)^T + \sum_{n=1}^{N^d} diag(\phi_n^d) \Bigg).\nonumber
\end{align}

We can find maximum likelihood estimates for the annotator biases $b^r$ by optimizing the lower bound on the marginal likelihood. The terms in  $\mathcal{L}$ that involve $b$ are
\begin{align}
\mathcal{L}_{[b]} &= \sum_{d=1}^D \sum_{r=1}^{R_d} \frac{p^r}{2} \bigg( 2 y^{d,r} b^r - 2 m^d b^r - (b^r)^2 \bigg).
\end{align}
Taking derivatives \mbox{w.r.t.} $b^r$ gives the following estimate for the bias of the $r^{th}$ annotator
\begin{align}
b^r = \frac{1}{D_r} \sum_{d=1}^{D_r} \Big( y^{d,r} - m^d \Big).
\end{align}

Similarly, we can find maximum likelihood estimates for the precisions $p^r$ of the different annotators by considering the terms in $\mathcal{L}$ that contain $p$
\begin{align}
\mathcal{L}_{[p]} = \sum_{d=1}^D \sum_{r=1}^{R_d} \bigg( \frac{1}{2} \log(p^r) - \frac{p^r v^d}{2} - \frac{p^r}{2} (y^{d,r} - m^d - b^r)^2 \bigg).
\end{align}
The maximum likelihood estimate for the precision (inverse variance) of the $r^{th}$ annotator is then given by
\begin{align}
p^r = \Bigg( \frac{1}{D_r} \sum_{d=1}^{D_r} \Big( v^d + ( y^{d,r} - m^d - b^r )^2 \Big) \Bigg)^{-1}.
\end{align}

Given a set of fitted parameters, it is then straightforward to make predictions for new documents: it is just necessary to infer the (approximate) posterior distribution over the word-topic assignments $z_n^d$ for all the words using the coordinates ascent updates of standard LDA (Eqs.~\ref{eq:gamma_update} and \ref{eq:phi_update_lda}), and then use the mean topic assignments $\bar{\phi}^d$ to make predictions $x_*^d = \eta^T \bar{\phi}^d$.

\subsection{Stochastic variational inference}

As we did for the classification model from Section~\ref{sec:classification}, we can envision developing a stochastic variational inference for the proposed regression model. In this case, the only ``global" latent variables are the per-topic distributions over words $\beta_k$. As for the ``local" latent variables, instead of a single variable $\lambda^d$, we now have two variables per-document: $m^d$ and $v^d$. The stochastic variational inference can then be summarized as shown in Algorithm~\ref{alg:svi_reg}. 
For added efficiency, one can also perform stochastic updates of the annotators biases $b^r$ and precisions $p^r$, by taking a step in the direction of the gradient of the noisy evidence lower bound scaled by the step-size $\rho_t$. 

\begin{algorithm}[t]
\caption{Stochastic variational inference for the proposed regression model}\label{alg:svi_reg}
\begin{algorithmic}[1]
\STATE Initialize $\bs\gamma^{(0)}$, $\bs\phi_{1:D}^{(0)}$, $\textbf{m}^{(0)}$, $\textbf{v}^{(0)}$, $\bs\zeta^{(0)}$, $\bs\xi^{(0)}_{1:R}$, $t=0$
\REPEAT
 \STATE Set t = t + 1
 \STATE Sample a document $\bw^d$ uniformly from the corpus
 \REPEAT
 \STATE Compute $\phi_n^d$ using \mbox{Eq.}~\ref{eq:phi_update_reg}, for $n \in \{1..N_d\}$
 \STATE Compute $\gamma^d$ using \mbox{Eq.}~\ref{eq:gamma_update}
 \STATE Compute $m^d$ using \mbox{Eq.}~\ref{eq:m_update}
 \STATE Compute $v^d$ using \mbox{Eq.}~\ref{eq:v_update}
 \UNTIL local parameters $\phi_n^d$, $\gamma^d$ and $\lambda^d$ converge
 \STATE Compute step-size $\rho_t = (t + delay)^{-\kappa}$
 \STATE Update topics variational parameters
 \begin{align}
 {\zeta_{i,j}}^{(t)} &= (1-\rho_t) \zeta_{i,j}^{(t-1)} + \rho_t \bigg(\tau + D \sum_{n=1}^{N_d} w_{n,j}^d  \phi_{n,i}^d\bigg)\nonumber
 \end{align}
\UNTIL global convergence criterion is met
\end{algorithmic}
\end{algorithm}

\section{Experiments}
\label{sec:experiments}

In this section, the proposed multi-annotator supervised LDA models for classification and regression (MA-sLDAc and MA-sLDAr, respectively) are validated using both simulated annotators on popular corpora and using real multiple-annotator labels obtained from Amazon Mechanical Turk.\footnote{Source code and datasets used are available at:\\ http://www.fprodrigues.com/software/} Namely, we shall consider the following real-world problems: classifying posts and news stories; classifying images according to their content; predicting number of stars that a given user gave to a restaurant based on the review; predicting movie ratings using the text of the reviews.

\subsection{Classification}
\label{subsec:classfication}

\subsubsection{Simulated annotators}

In order to first validate the proposed model for classification problems in a slightly more controlled environment, the well-known 20-Newsgroups benchmark corpus \cite{Lang95} was used by simulating multiple annotators with different levels of expertise. The 20-Newsgroups consists of twenty thousand messages taken from twenty newsgroups, and is divided in six super-classes, which are, in turn, partitioned in several sub-classes. For this first set of experiments, only the four most populated super-classes were used: ``computers", ``science", ``politics" and ``recreative". The preprocessing of the documents consisted of stemming and stop-words removal. After that, 75\% of the documents were randomly selected for training and the remaining 25\% for testing. 

The different annotators were simulated by sampling their answers from a multinomial distribution, where the parameters are given by the lines of the annotators' confusion matrices. Hence, for each annotator $r$, we start by pre-defining a confusion matrix $\bs\pi^r$ with elements $\pi_{c,l}^r$, which correspond to the probability that the annotators' answer is $l$ given that the true label is $c$, $p(y^r = l | c)$. Then, the answers are sampled i.i.d. from $y^r \sim \mbox{Mult}(y^r|\pi_{c,l}^r)$. This procedure was used to simulate 5 different annotators with the following accuracies: 0.737, 0.468, 0.284, 0.278, 0.260. In this experiment, no repeated labelling was used. Hence, each annotator only labels roughly one-fifth of the data. When compared to the ground truth, the simulated answers revealed an accuracy of 0.405. See Table~\ref{table:data_stats} for an overview of the details of the classification datasets used. 

\begin{table*}[t!]
\caption{Overall statistics of the classification datasets used in the experiments.}
\label{table:data_stats}
\begin{center}
\begin{tabular}{c|c|c|c|c|c|c}
Dataset&\specialcell{Num.\\classes} & \specialcell{Train/test\\sizes} & \specialcell{Annotators\\source}  & \specialcell{Num. answers per\\ instance ($\pm$ stddev.)}  & \specialcell{Mean annotators\\ accuracy ($\pm$ stddev.)}  & \specialcell{Maj. vot.\\ accuracy}  \\
\hline
20 Newsgroups & 4 & 11536/3846 & Simulated & 1.000 $\pm$ 0.000 & 0.405 $\pm$ 0.182 & 0.405 \\
Reuters-21578 & 8 & 1800/5216 & Mech. Turk & 3.007 $\pm$ 1.019 & 0.568 $\pm$ 0.262 & 0.710 \\
LabelMe & 8 & 1000/1688 & Mech. Turk & 2.547 $\pm$ 0.576 & 0.692 $\pm$ 0.181 & 0.769 \\
\end{tabular}
\end{center}
\vspace*{-0.3cm}
\end{table*}%

Both the \textit{batch} and the stochastic variational inference (\textit{svi}) versions of the proposed model (MA-sLDAc) are compared with the following baselines:

\begin{itemize}[itemsep=0.02cm]
\item \textit{LDA + LogReg (mv)}: This baseline corresponds to applying unsupervised LDA to the data, and learning a logistic regression classifier on the inferred topics distributions of the documents. The labels from the different annotators were aggregated using majority voting (mv). Notice that, when there is a single annotator label per instance, majority voting is equivalent to using that label for training. This is the case of the 20-Newsgroups' simulated annotators, but the same does not apply for the experiments in Section~\ref{subsec:realannotators}.
\item \textit{LDA + Raykar}: For this baseline, the model of \cite{Raykar2010} was applied using the documents' topic distributions inferred by LDA as features.
\item \textit{LDA + Rodrigues}: This baseline is similar to the previous one, but uses the model of \cite{Rodrigues2013a} instead. 
\item \textit{Blei 2003 (mv)}: The idea of this baseline is to replicate a popular state-of-the-art approach for document classification. Hence, the approach of \cite{Blei2003} was used. It consists of applying LDA to extract the documents' topics distributions, which are then used to train a SVM. Similarly to the previous approach, the labels from the different annotators were aggregated using majority voting (mv).
\item \textit{sLDA (mv)}: This corresponds to using the classification version of sLDA \cite{Wang2009} with the labels obtained by performing majority voting (mv) on the annotators' answers.
\end{itemize}

\begin{figure}[!t]
\includegraphics[trim=0 1.1cm 0 0.5cm, clip, scale=.34]{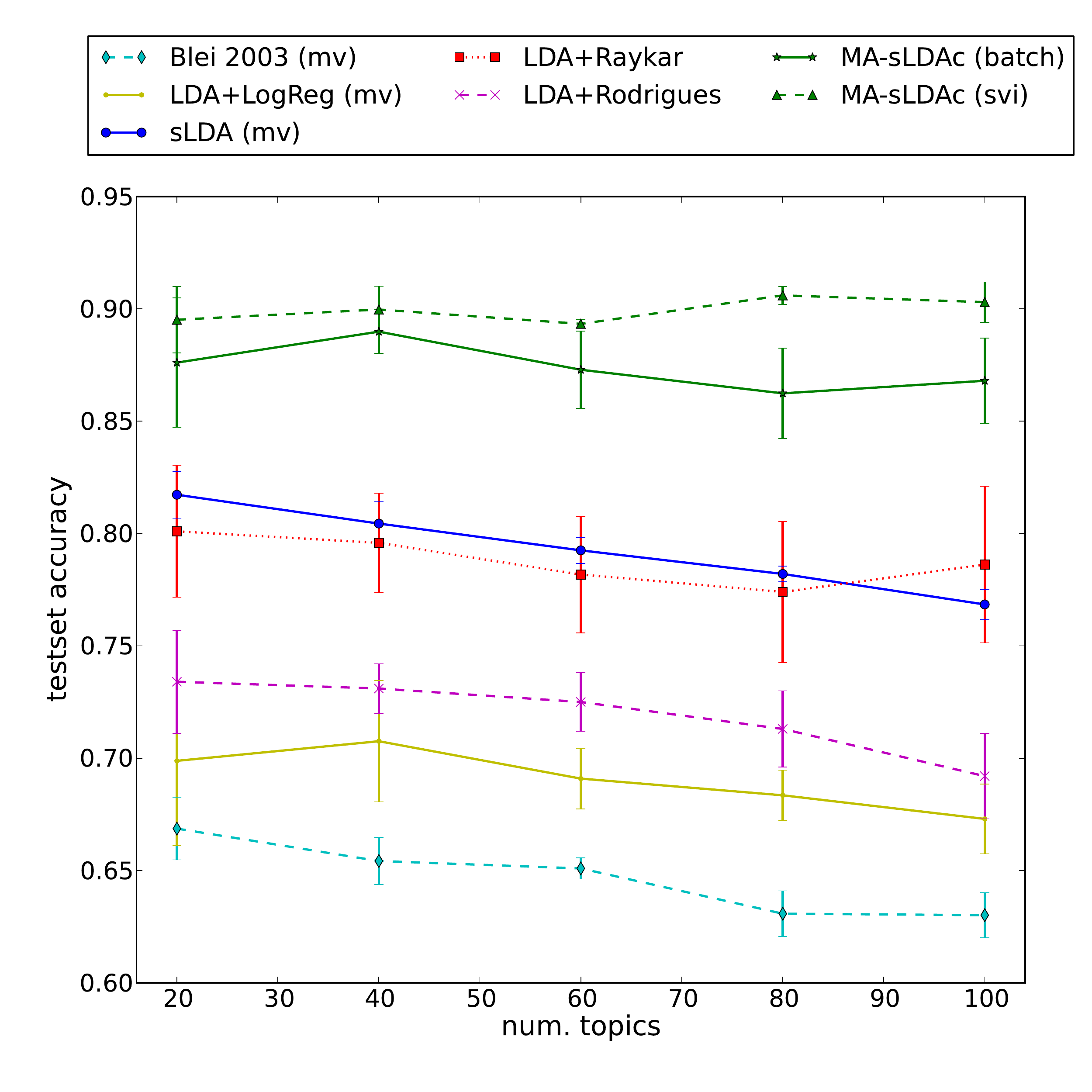}
\caption{Average testset accuracy (over 5 runs; $\pm$ stddev.) of the different approaches on the 20-Newsgroups data.}
\label{fig:20newsgroups}
\end{figure}

For all the experiments the hyper-parameters $\alpha$, $\tau$ and $\omega$ were set using a simple grid search in the collection $\{0.01,0.1,1.0,10.0\}$. The same approach was used to optimize the hyper-parameters of the all the baselines. For the \textit{svi} algorithm, different mini-batch sizes and forgetting rates $\kappa$ were tested. For the 20-Newsgroup dataset, the best results were obtained with a mini-batch size of 500 and $\kappa = 0.6$. The $delay$ was kept at 1. The results are shown in Fig.~\ref{fig:20newsgroups} for different numbers of topics, where we can see that the proposed model outperforms all the baselines, being the \textit{svi} version the one that performs best. 

\begin{figure}[!t]
\centering
\includegraphics[trim=0 0.3cm 0 0, clip, scale=.4]{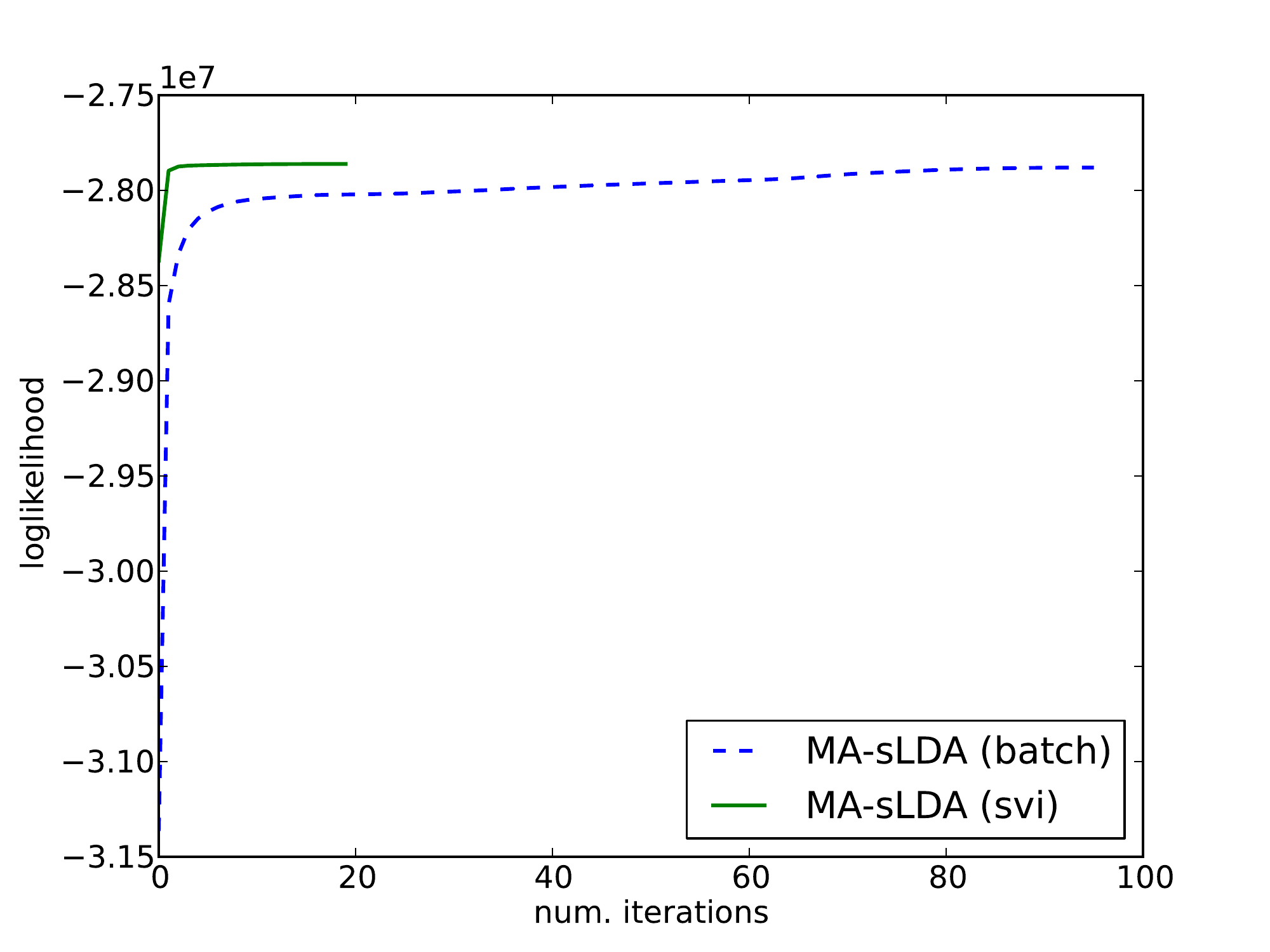}
\caption{Comparison of the log marginal likelihood between the \textit{batch} and the stochastic variational inference (\textit{svi}) algorithms on the 20-Newsgroups corpus.}
\label{fig:likelihoods}
\end{figure}

In order to assess the computational advantages of the stochastic variational inference (\textit{svi}) over the \textit{batch} algorithm, the log marginal likelihood (or log evidence) was plotted against the number of iterations. Fig.~\ref{fig:likelihoods} shows this comparison. Not surprisingly, the \textit{svi} version converges much faster to higher values of the log marginal likelihood when compared to the \textit{batch} version, which reflects the efficiency of the \textit{svi} algorithm. 

\subsubsection{Amazon Mechanical Turk}
\label{subsec:realannotators}

In order to validate the proposed classification model in real crowdsourcing settings, Amazon Mechanical Turk (AMT) was used to obtain labels from multiple annotators for two popular datasets: Reuters-21578 \cite{Lewis1997} and LabelMe \cite{Russell2008}. 

\begin{figure}[t]
\centering
\includegraphics[trim=0 0.9cm 0 0, clip, width=8.5cm]{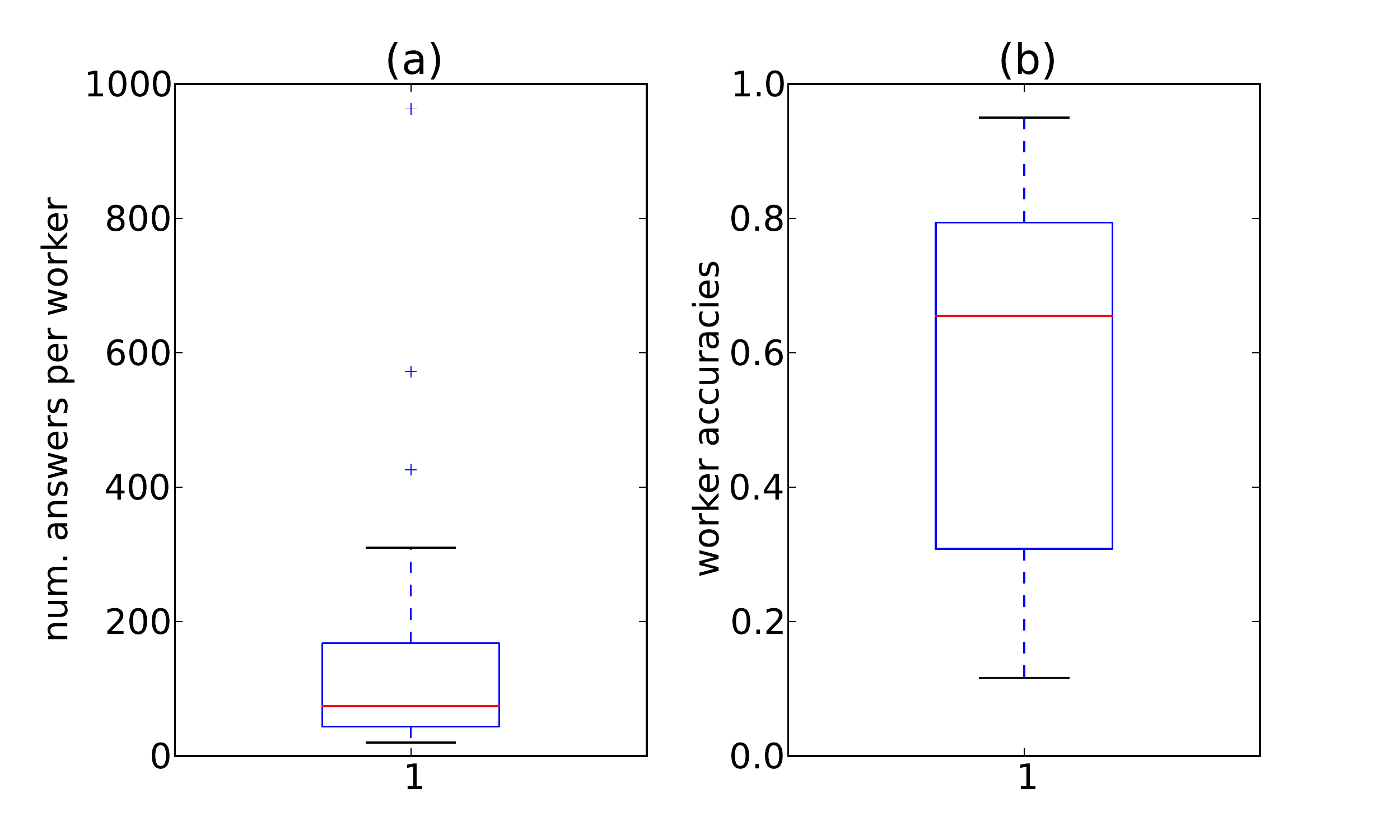}
\caption{Boxplot of the number of answers per worker (a) and their respective accuracies (b) for the Reuters dataset.}
\label{fig:boxplot_reuters}
\end{figure}

The Reuters-21578 is a collection of manually categorized newswire stories with labels such as Acquisitions, Crude-oil, Earnings or Grain. For this experiment, only the documents belonging to the ModApte split were considered with the additional constraint that the documents should have no more than one label. This resulted in a total of 7016 documents distributed among 8 classes. Of these, 1800 documents were submitted to AMT for multiple annotators to label, giving an average of approximately 3 answers per document (see Table~\ref{table:data_stats} for further details). The remaining 5216 documents were used for testing. The collected answers yield an average worker accuracy of 56.8\%. Applying majority voting to these answers reveals a ground truth accuracy of 71.0\%. Fig.~\ref{fig:boxplot_reuters} shows the boxplots of the number of answers per worker and their accuracies. Observe how applying majority voting yields a higher accuracy than the median accuracy of the workers. 

\begin{figure}[!t]
\includegraphics[trim=0 1.1cm 0 0.5cm, clip, scale=.34]{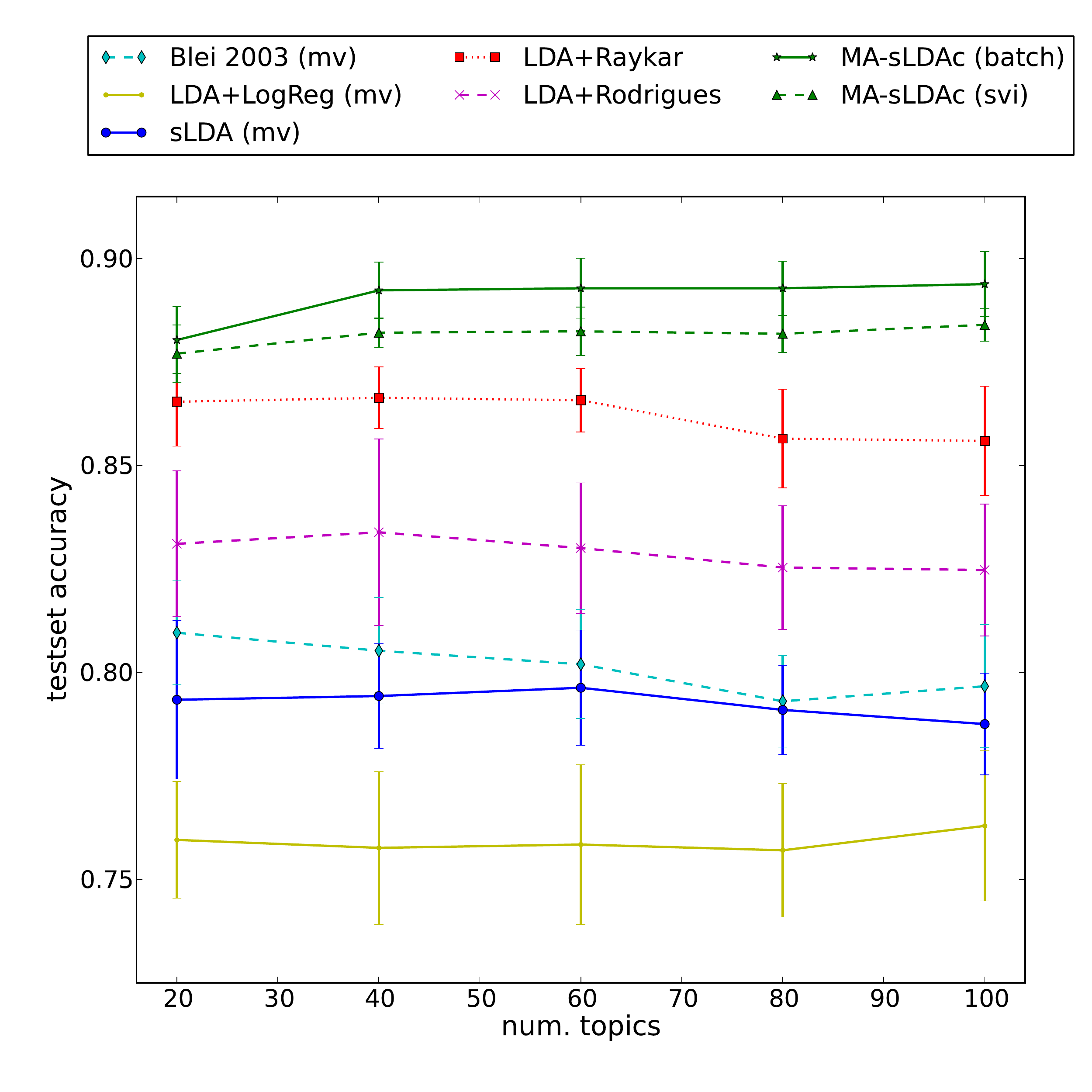}
\caption{Average testset accuracy (over 30 runs; $\pm$ stddev.) of the different approaches on the Reuters data.}
\label{fig:reuters}
\end{figure}

The results obtained by the different approaches are given in Fig.~\ref{fig:reuters}, where it can be seen that the proposed model (MA-sLDAc) outperforms all the other approaches. For this dataset, the \textit{svi} algorithm is using mini-batches of 300 documents. 

\begin{figure}[t]
\centering
\includegraphics[trim=0 0.9cm 0 0, clip, width=8.5cm]{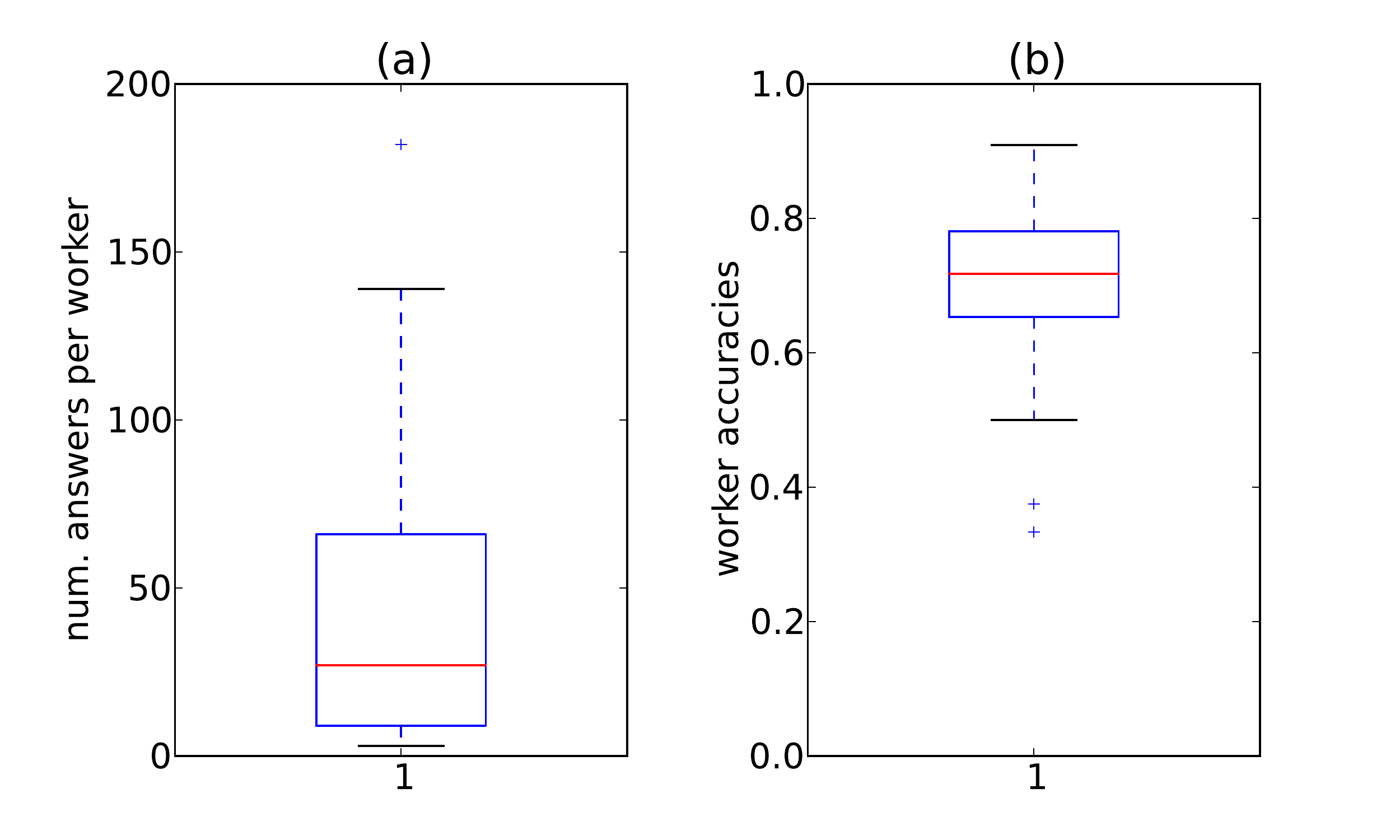}
\caption{Boxplot of the number of answers per worker (a) and their respective accuracies (b) for the LabelMe dataset.}
\label{fig:boxplot_labelme}
\end{figure}

The proposed model was also validated using a dataset from the computer vision domain: LabelMe \cite{Russell2008}. In contrast to the Reuters and Newsgroups corpora, LabelMe is an open online tool to annotate images. Hence, this experiment allows us to see how the proposed model generalizes beyond non-textual data. Using the Matlab interface provided in the projects' website, we extracted a subset of the LabelMe data, consisting of all the 256 x 256 images with the categories: ``highway", ``inside city", ``tall building", ``street", ``forest", ``coast", ``mountain" or ``open country". This allowed us to collect a total of 2688 labeled images. Of these, 1000 images were given to AMT workers to classify with one of the classes above. Each image was labeled by an average of 2.547 workers, with a mean accuracy of 69.2\%. When majority voting is applied to the collected answers, a ground truth accuracy of 76.9\% is obtained. Fig.~\ref{fig:boxplot_labelme} shows the boxplots of the number of answers per worker and their accuracies. Interestingly, the worker accuracies are much higher and their distribution is much more concentrated than on the Reuters-21578 data (see Fig.~\ref{fig:boxplot_reuters}), which suggests that this is an easier task for the AMT workers. 

The preprocessing of the images used is similar to the approach in \cite{FeiFei2005}. It uses 128-dimensional SIFT \cite{Lowe1999} region descriptors selected by a sliding grid spaced at one pixel. This sliding grid extracts local regions of the image with sizes uniformly sampled between 16 x 16 and 32 x 32 pixels. The 128-dimensional SIFT descriptors produced by the sliding window are then fed to a k-means algorithm (with k=200) in order construct a vocabulary of 200 ``visual words". This allows us to represent the images with a bag of visual words model.

\begin{figure}[!t]
\includegraphics[trim=0 1.1cm 0 0.5cm, clip, scale=.34]{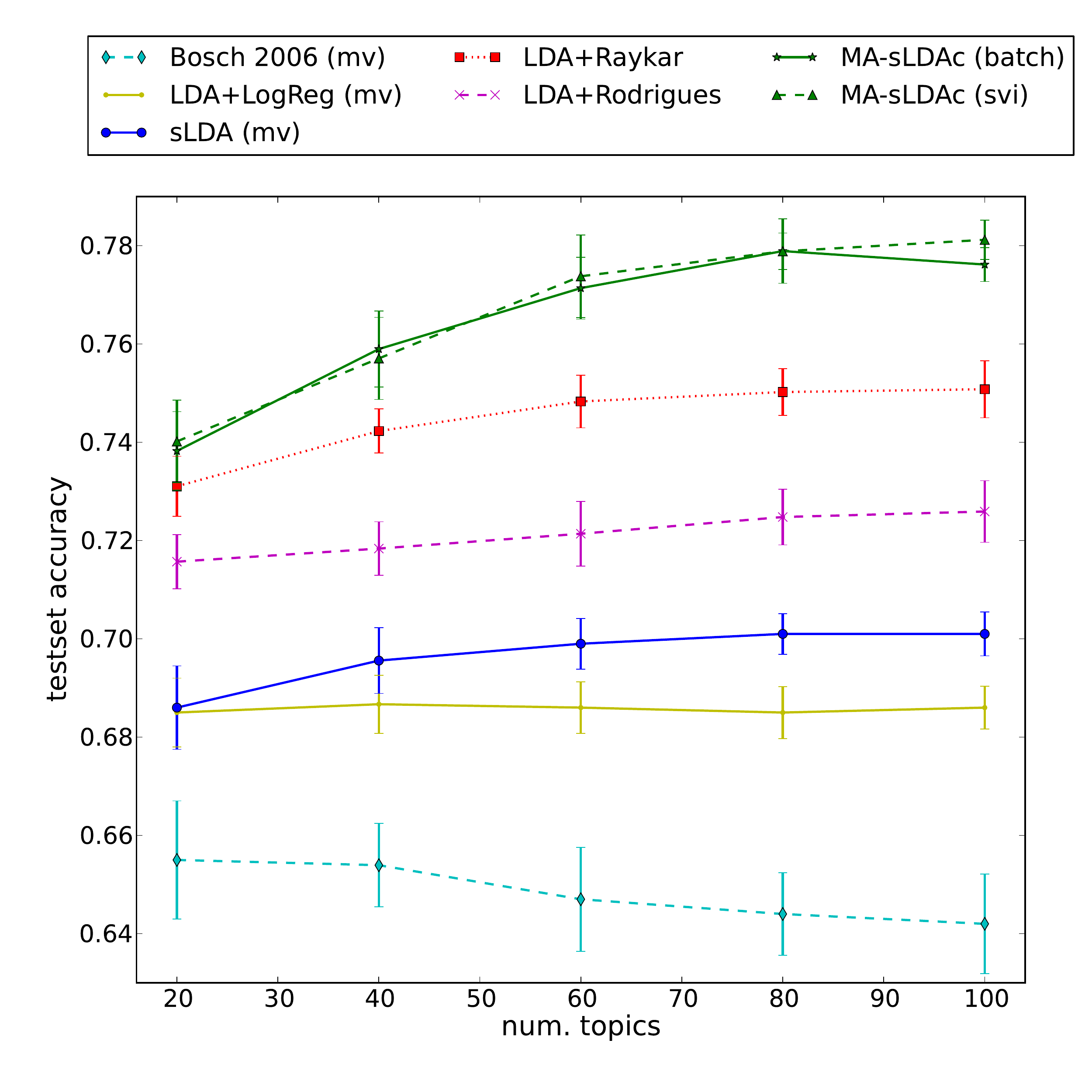}
\caption{Average testset accuracy (over 30 runs; $\pm$ stddev.) of the different approaches on the LabelMe data.}
\label{fig:labelme}
\vspace*{-0.2cm}
\end{figure}

With the purpose of comparing the proposed model with a popular state-of-the-art approach for image classification, for the LabelMe dataset, the following baseline was introduced:
\begin{itemize}
\item \textit{Bosch 2006 (mv)}: This baseline is similar to one in \cite{Bosch2006}. The authors propose the use of pLSA to extract the latent topics, and the use of k-nearest neighbor (kNN) classifier using the documents' topics distributions. For this baseline, unsupervised LDA is used instead of pLSA, and the labels from the different annotators for kNN (with $k=10$) are aggregated using majority voting (mv).
\end{itemize}
The results obtained by the different approaches for the LabelMe data are shown in Fig.~\ref{fig:labelme}, where the \textit{svi} version is using mini-batches of 200 documents. 

\begin{figure*}[!t]
\centering
\hspace{-0.5cm}
\subfloat[annotator 1]{\includegraphics[width=0.365\linewidth]{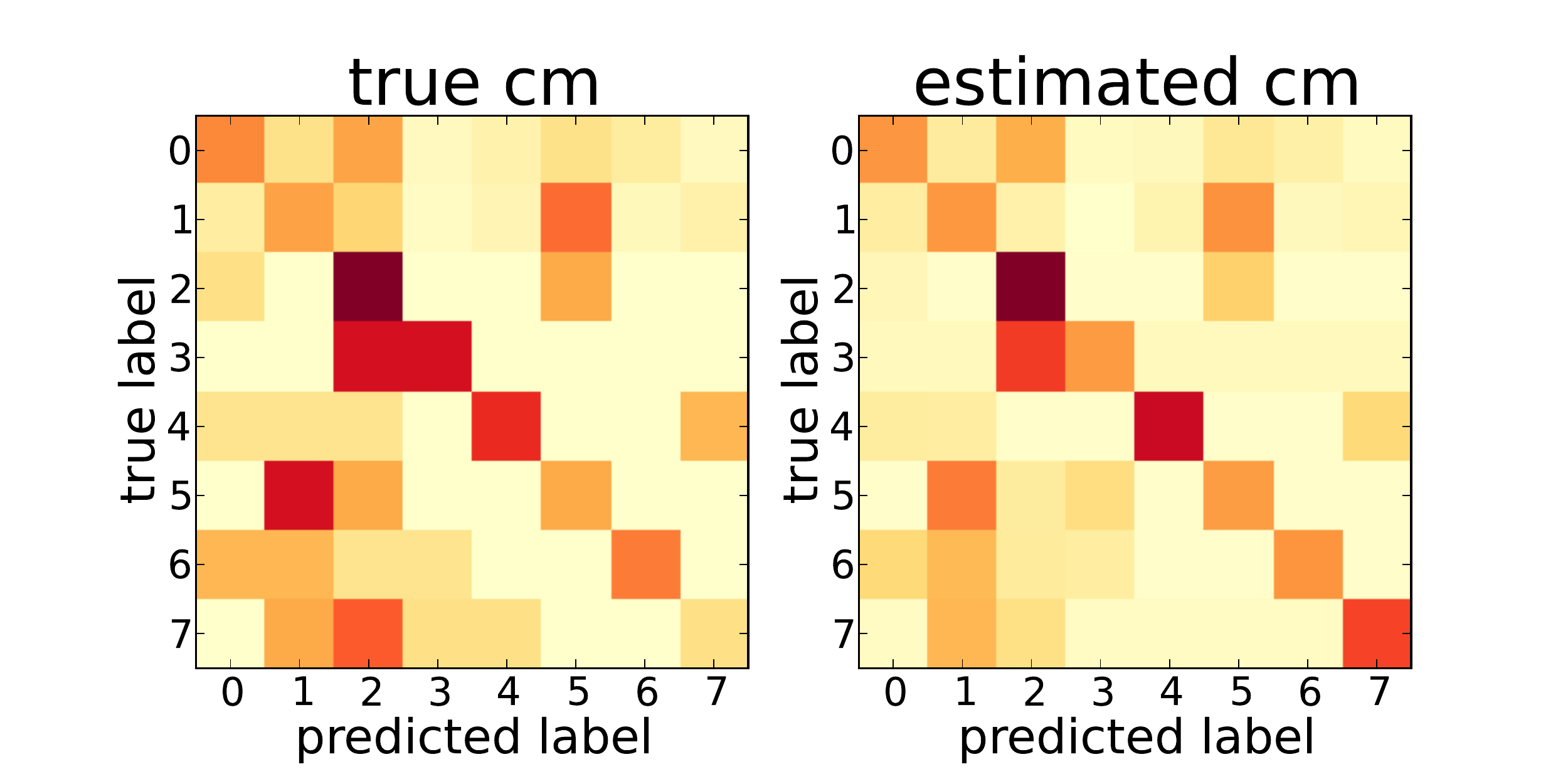}}\hspace{-0.75cm}
\subfloat[annotator 2]{\includegraphics[width=0.365\linewidth]{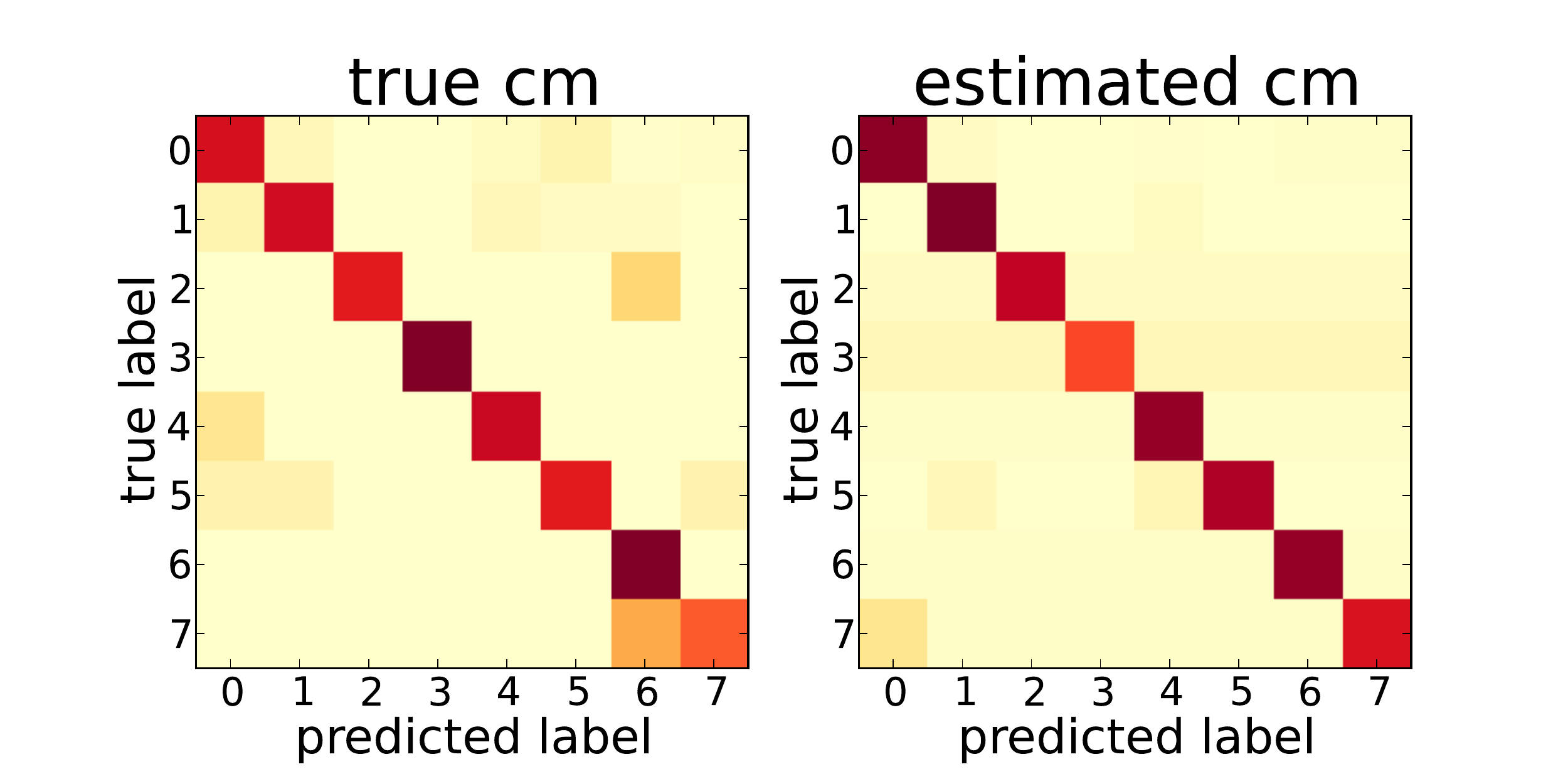}}\hspace{-0.75cm}
\subfloat[annotator 3]{\includegraphics[width=0.365\linewidth]{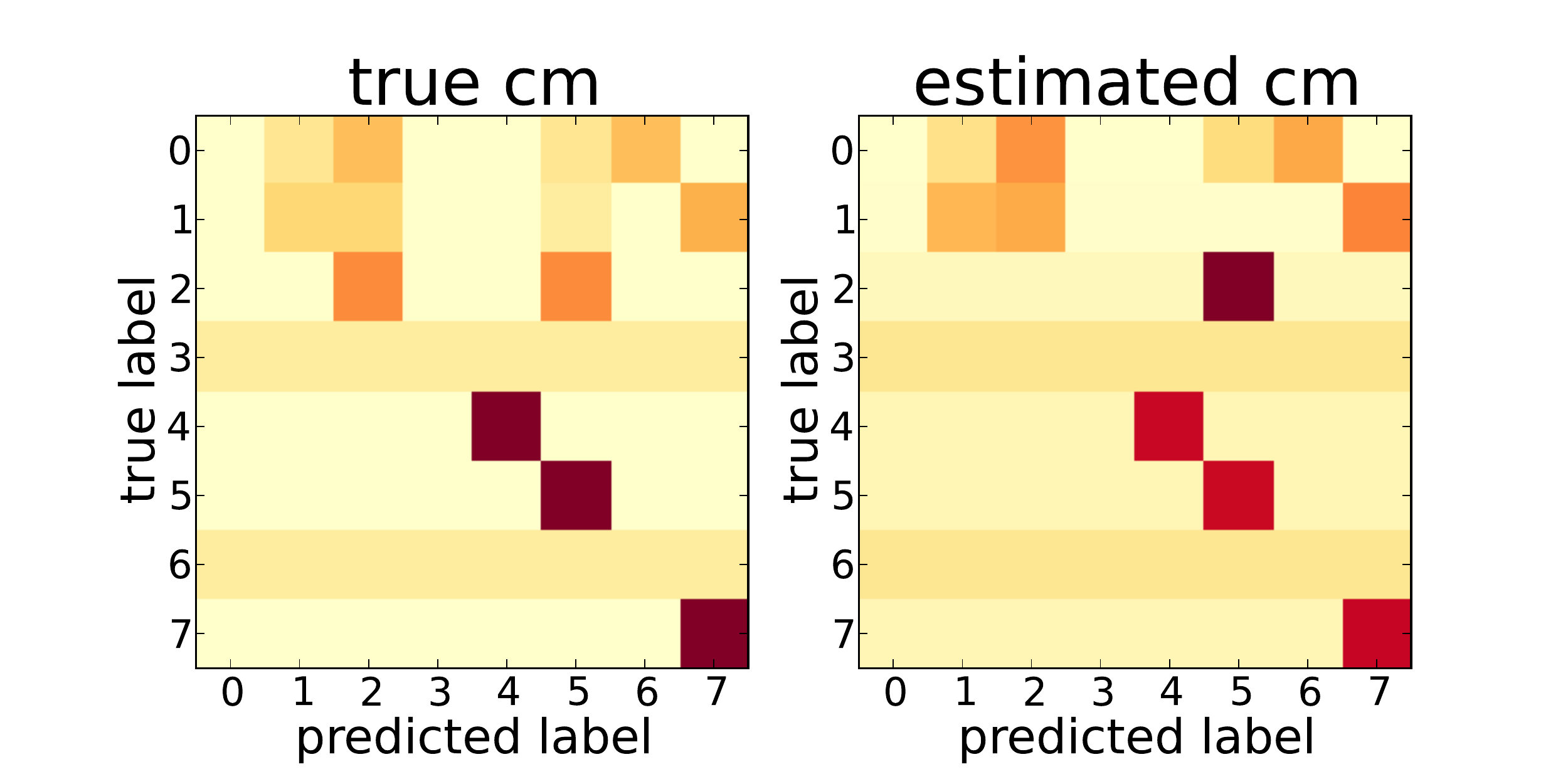}}%

\hspace{-0.5cm}
\subfloat[annotator 4]{\includegraphics[width=0.365\linewidth]{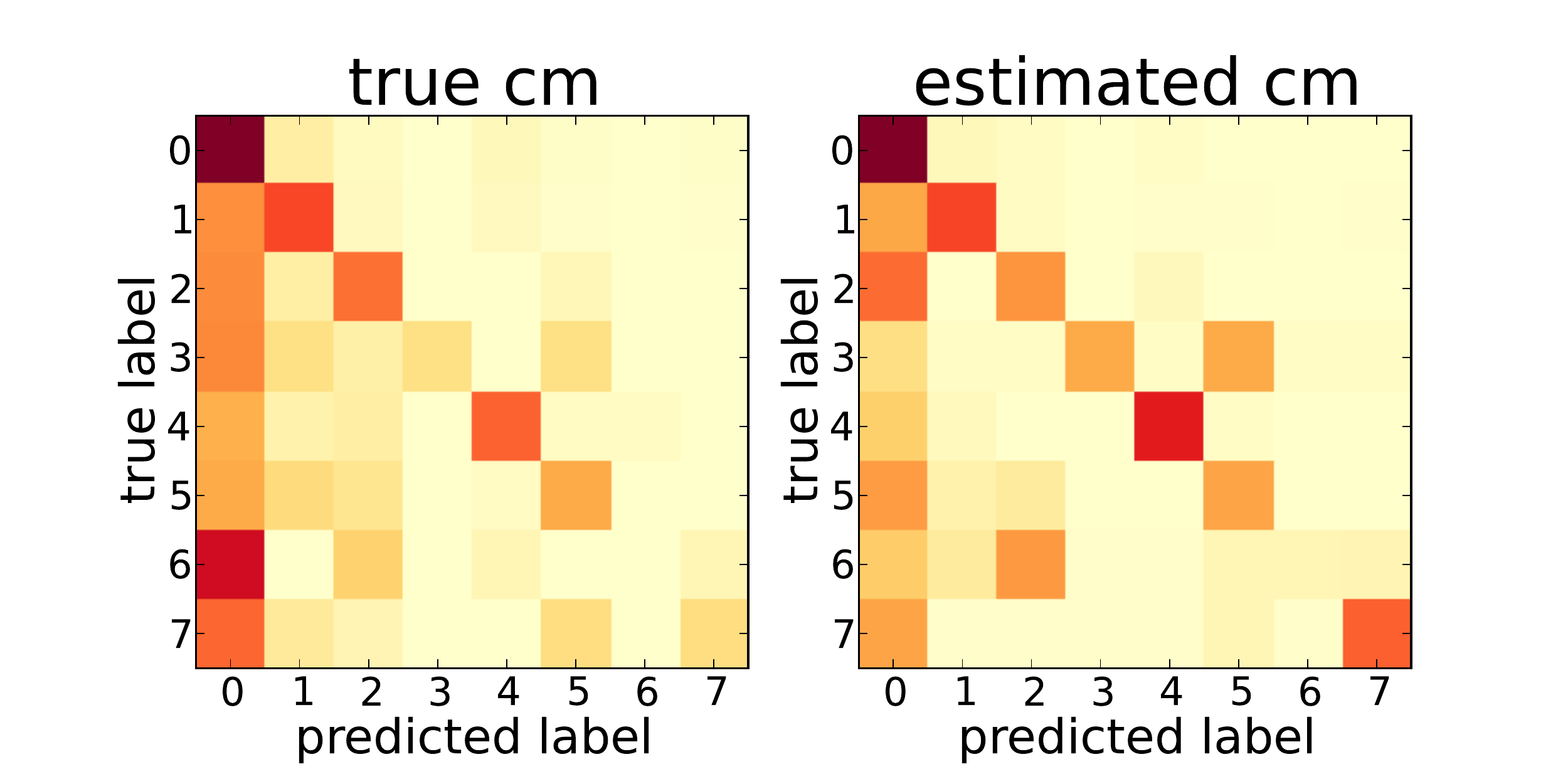}}\hspace{-0.75cm}
\subfloat[annotator 5]{\includegraphics[width=0.365\linewidth]{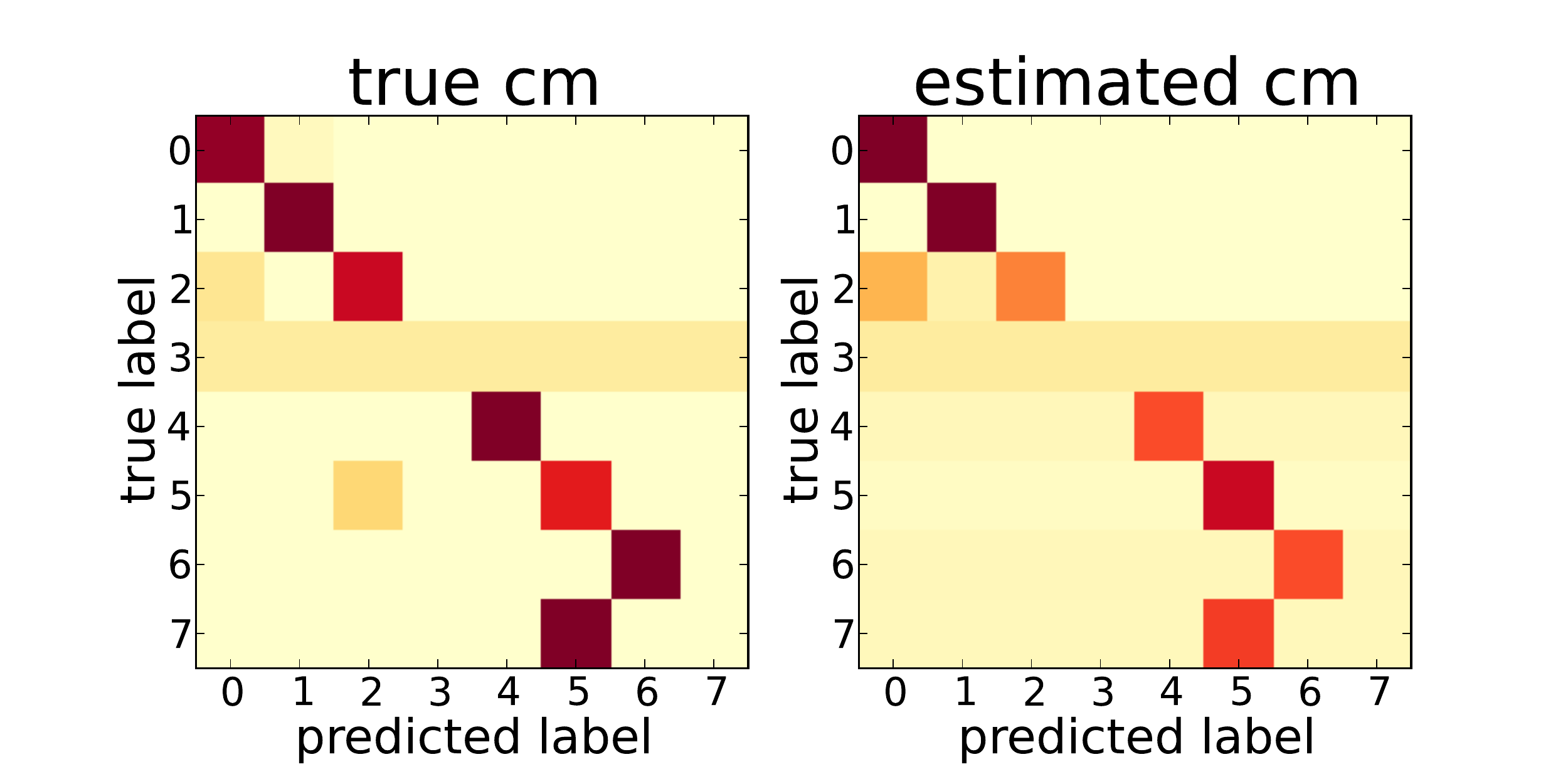}}\hspace{-0.75cm}
\subfloat[annotator 6]{\includegraphics[width=0.365\linewidth]{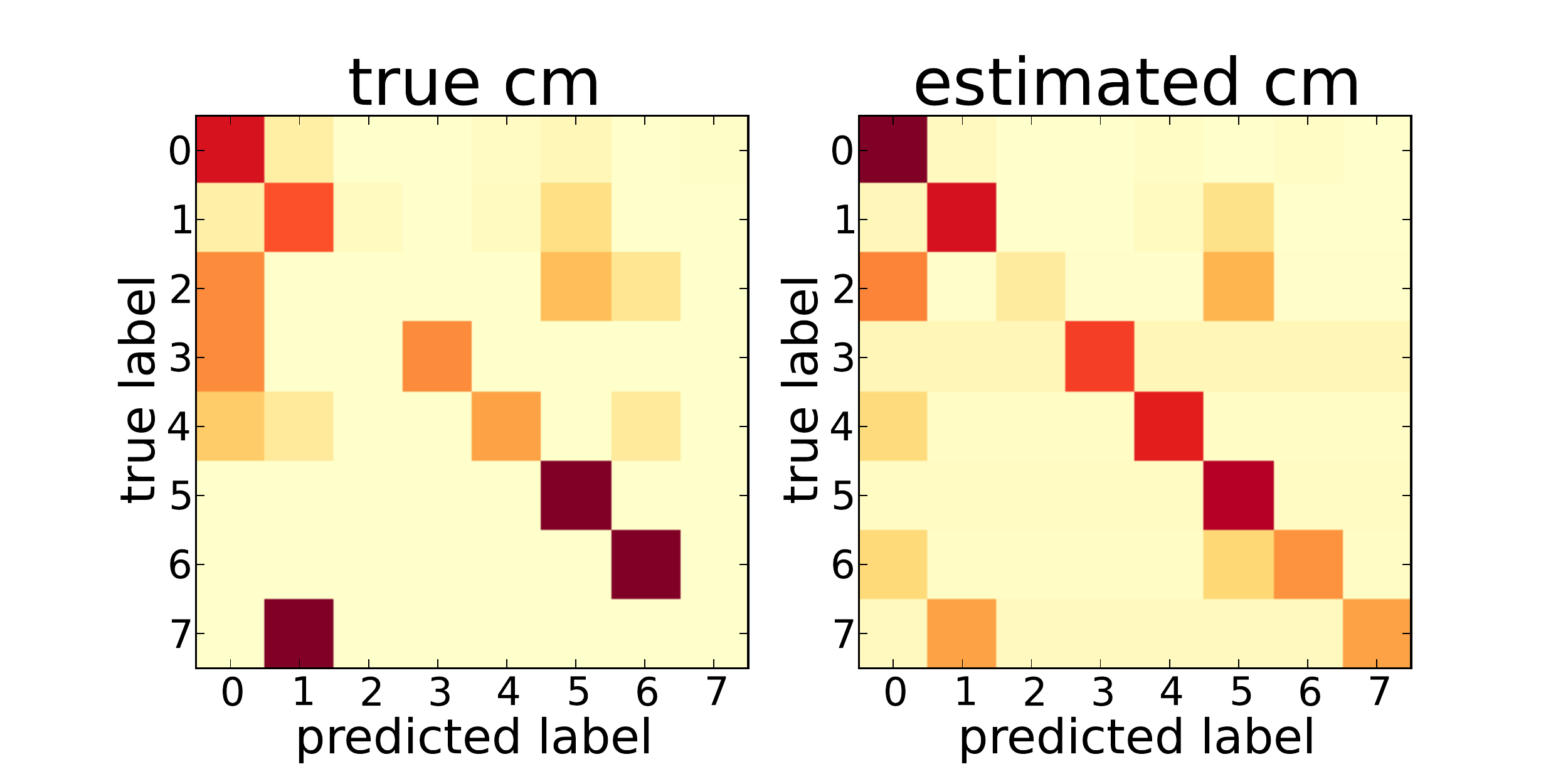}}%
\caption{True vs. estimated confusion matrix (cm) of 6 different workers of the Reuters-21578 dataset.}
\label{fig:conf_mats_reuters}
\end{figure*}

\begin{table}[htp]
\caption{Results for 4 example LabelMe images.}
\begin{center}
\begin{tabular}{l | c }
\includegraphics[trim=0 0 0 -0.2cm, scale=.3]{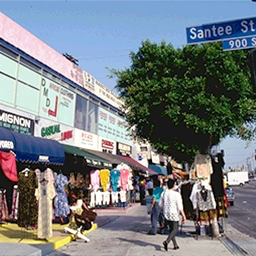} & \multirow{3}{*}[8.5em]{\includegraphics[trim={0.2cm 1.6cm 0.4cm 0.7cm},clip,scale=.5]{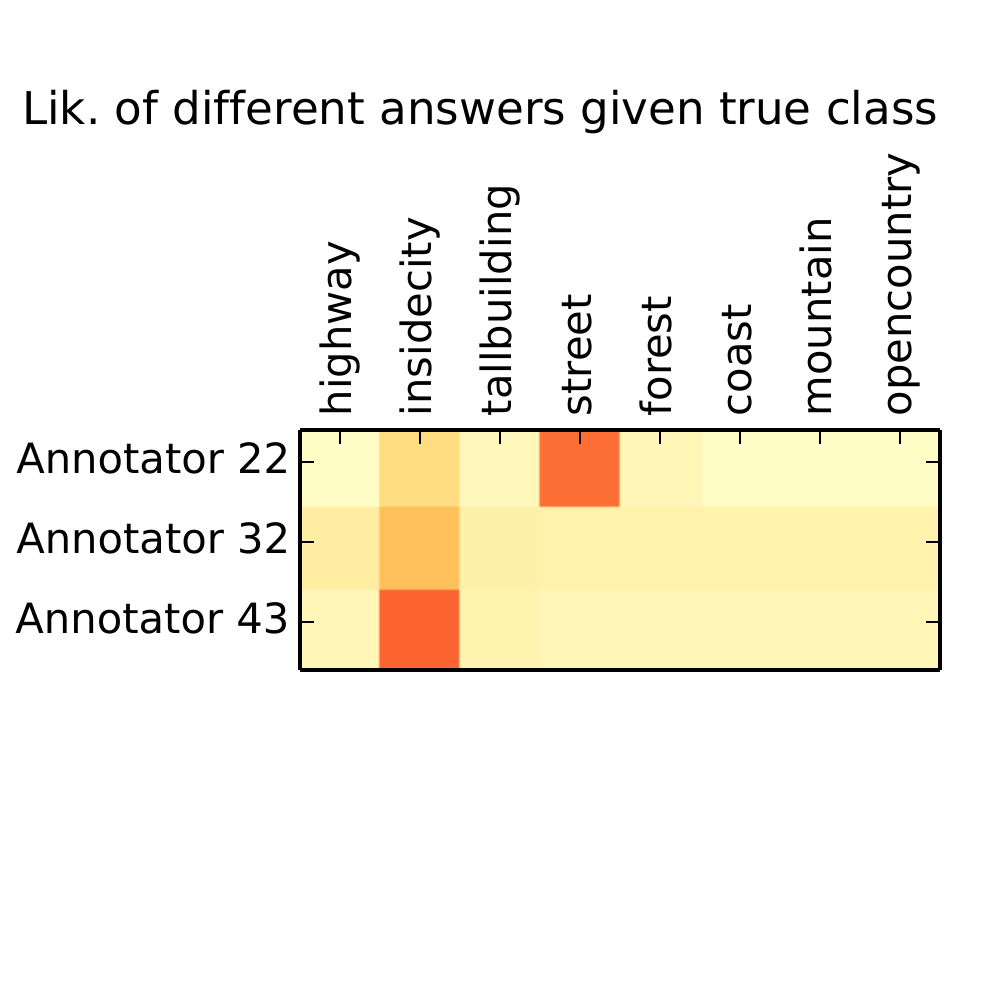}} \\
True label: street & \\
Annotator 22: street & \\
Annotator 32: inside city & Inferred ground truth: street \\
Annotator 43: inside city & \\
\hline
\includegraphics[trim=0 0 0 -0.2cm, scale=.3]{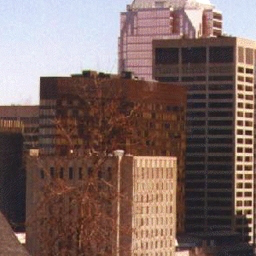} & \multirow{3}{*}[8.5em]{\includegraphics[trim={0.2cm 1.6cm 0.4cm 0.7cm},clip,scale=.5]{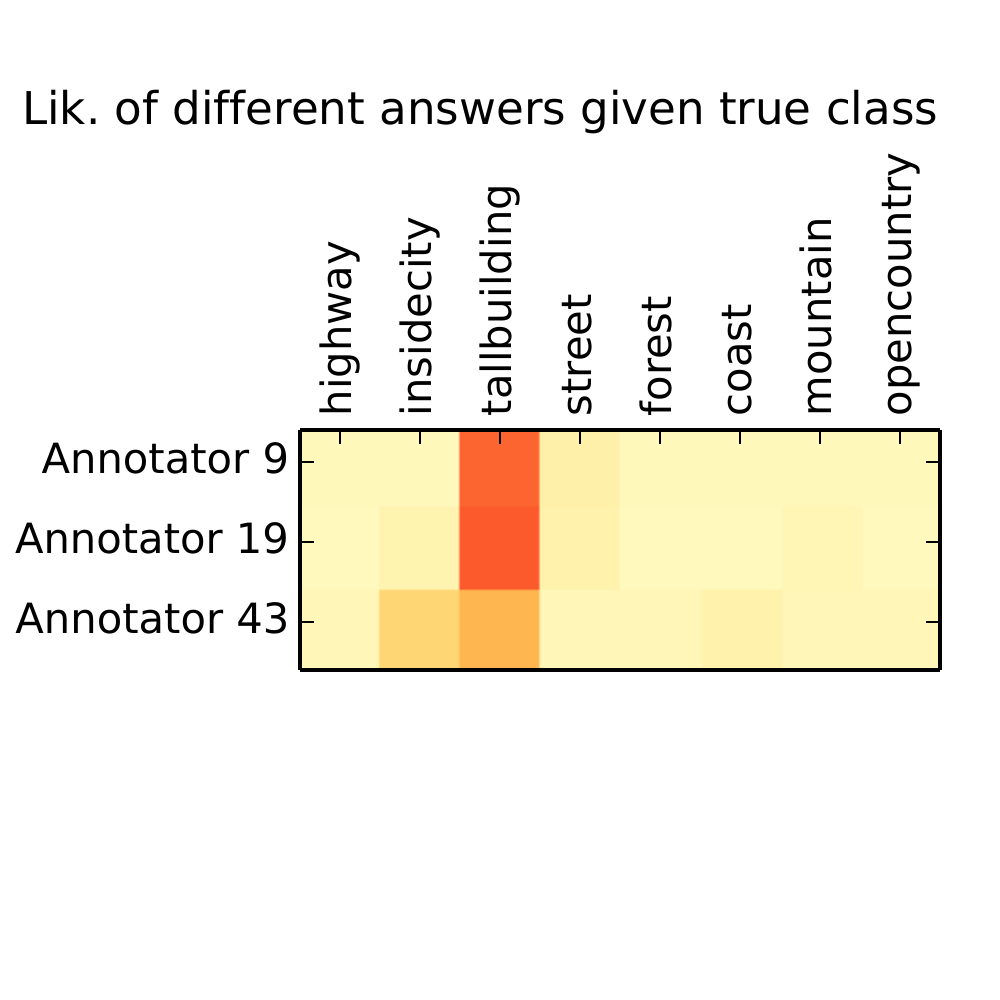}} \\
True label: tall building & \\
Annotator 9: tall building & \\
Annotator 19: street & Inferred ground truth: tall building \\
Annotator 43: inside city & \\
\hline
\includegraphics[trim=0 0 0 -0.2cm,scale=.3]{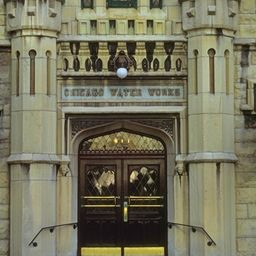} & \multirow{3}{*}[8.5em]{\includegraphics[trim={0.2cm 1.6cm 0.4cm 0.7cm},clip,scale=.5]{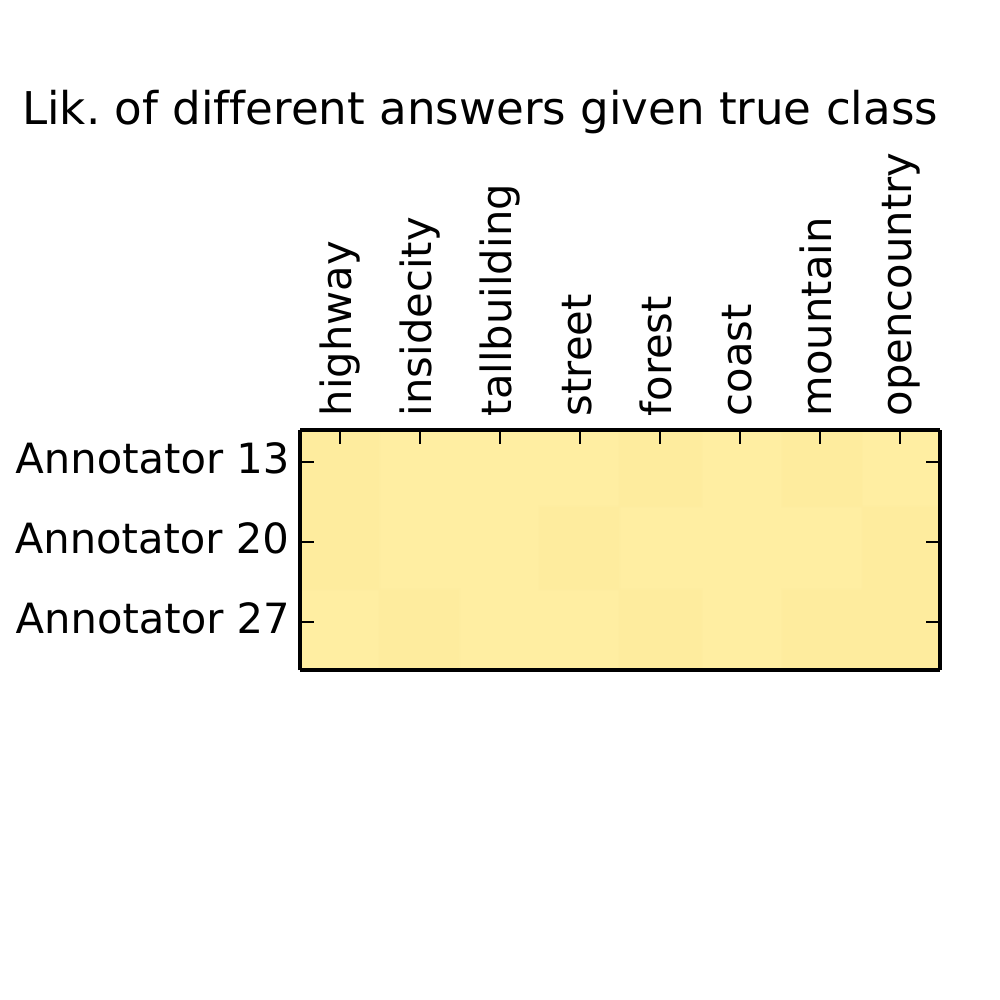}} \\
True label: inside city & \\
Annotator 13: inside city & \\
Annotator 20: tall building & Inferred ground truth: tall building \\
Annotator 27: tall building & \\
\hline
\includegraphics[trim=0 0 0 -0.2cm, scale=.3]{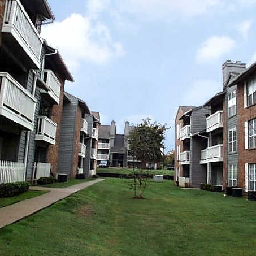} & \multirow{3}{*}[8.5em]{\includegraphics[trim={0.2cm 1.6cm 0.4cm 0.7cm},clip,scale=.5]{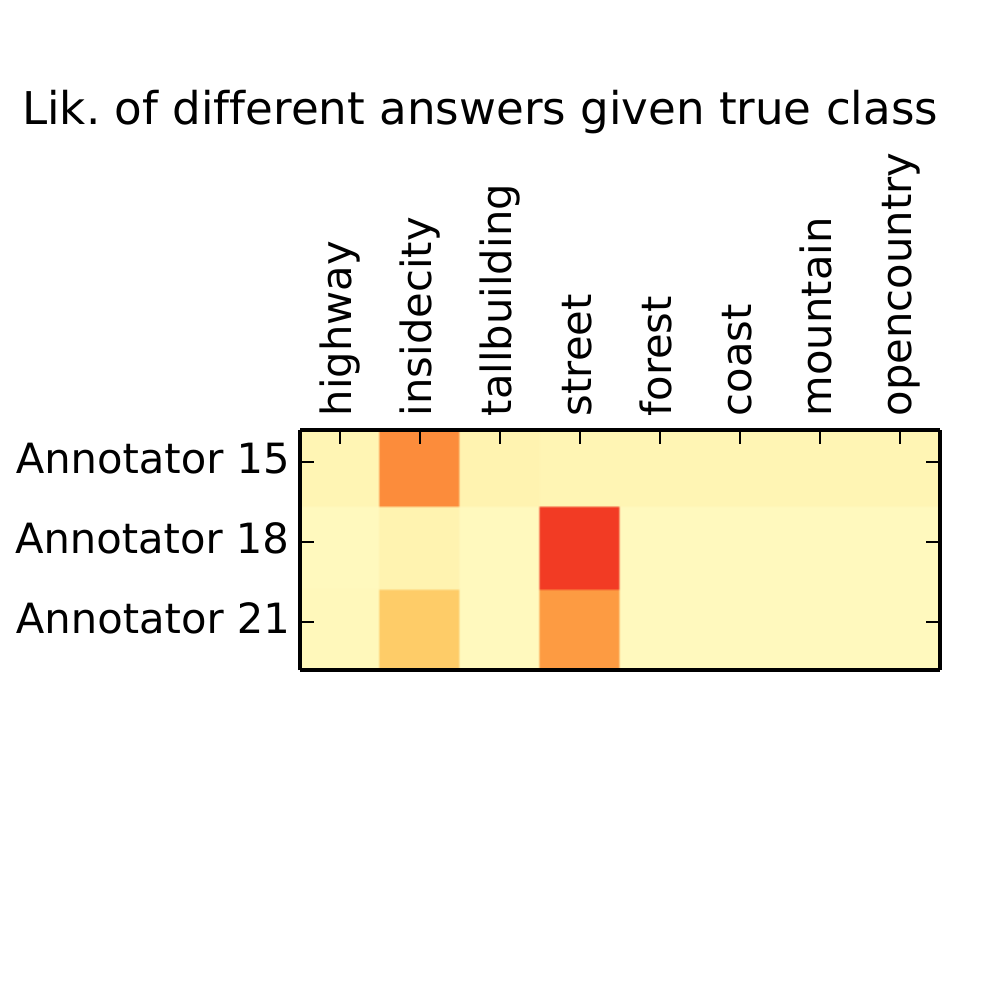}} \\
True label: street & \\
Annotator 15: inside city & \\
Annotator 18: inside city & Inferred ground truth: street \\
Annotator 21: inside city & \\
\end{tabular}
\end{center}
\label{fig:examples_images}
\end{table}%

Analyzing the results for the Reuters-21578 and LabelMe data, we can observe that MA-sLDAc outperforms all the baselines, with slightly better accuracies for the \textit{batch} version, especially in the Reuters data. Interestingly, the second best results are consistently obtained by the multi-annotator approaches, which highlights the need for accounting for the noise and biases of the answers of the different annotators. 

In order to verify that the proposed model was estimating the (normalized) confusion matrices $\bs\pi^r$ of the different workers correctly, a random sample of them was plotted against the true confusion matrices (\mbox{i.e.} the normalized confusion matrices evaluated against the true labels). Figure~\ref{fig:conf_mats_reuters} shows the results obtained with 60 topics on the Reuters-21578 dataset, where the color intensity of the cells increases with the magnitude of the value of $p(y^{d,r} = l | c^d) = \pi_{c,l}^r$ (the supplementary material provides a similar figure for the LabelMe dataset). Using this visualization we can verify that the AMT workers are quite heterogeneous in their labeling styles and in the kind of mistakes they make, with several workers showing clear biases (\mbox{e.g.} workers 3 and 4), while others made mistakes more randomly (\mbox{e.g.} worker 1). Nevertheless, the proposed is able to capture these patterns correctly and account for effect. 

To gain further insights, Table~\ref{fig:examples_images} shows 4 example images from the LabelMe dataset, along with their true labels, the answers provided by the different workers, the true label inferred by the proposed model and the likelihood of the different possible answers given the true label for each annotator ($\pi_{c^d,l}^r$ for $l \in \{1,...,C\}$) using a color-coding scheme similar to Fig.~\ref{fig:conf_mats_reuters}. In the first example, although majority voting suggests ``inside city" to be the correct label, we can see that the model has learned that annotators 32 and 43 are very likely to provide the label ``inside city" when the true label is actually ``street", and it is able to leverage that fact to infer that the correct label is ``street". Similarly, in the second image the model is able to infer the correct true label from 3 conflicting labels. However, in the third image the model is not able to recover the correct true class, which can be explained by it not having enough evidence about the annotators and their reliabilities and biases (likelihood distribution for these cases is uniform). In fact, this raises interesting questions regarding requirements for the minimum number of labels per annotator, their reliabilities and their coherence. Finally, for the fourth image, somehow surprisingly, the model is able to infer the correct true class, even though all 3 annotators labeled it as ``inside city".

\subsection{Regression}
\label{subsec:regression}

\subsubsection{Simulated annotators}
\label{subsec:we8there}

As for proposed classification model, we start by validating MA-sLDAr using simulated annotators on a popular corpus where the documents have associated targets that we wish to predict. For this purpose, we shall consider a dataset of user-submitted restaurant reviews from the website we8there.com. This dataset was originally introduced in \cite{Maua2009} and it consists of 6260 reviews. For each review, there is a five-star rating on four specific aspects of quality (food, service, value, and atmosphere) as well as the overall experience. Our goal is then to predict the overall experience of the user based on his comments in the review. We apply the same preprocessing as in \cite{Taddy2013}, which consists in tokenizing the text into bigrams and discarding those that appear in less than ten reviews. The preprocessing of the documents consisted of stemming and stop-words removal. After that, 75\% of the documents were randomly selected for training and the remaining 25\% for testing. 

As with the classification model, we seek to simulate an heterogeneous set of annotators in terms of reliability and bias. Hence, in order to simulate an annotator $r$, we proceed as follows: let $x^d$ be the true review of the restaurant; we start by assigning a given bias $b^r$ and precision $p^r$ to the reviewers, depending on what type of annotator we wish to simulate (see Fig.~\ref{fig:annotators_example}); we then sample a simulated answer as $y^{d,r} \sim \N(x^d + b^r, 1/p^r)$. Using this procedure, we simulated 5 annotators with the following (bias, precision) pairs: (0.1, 10), (-0.3, 3), (-2.5, 10), (0.1, 0.5)  and (1, 0.25). The goal is to have 2 good annotators (low bias, high precision), 1 highly biased annotator and 2 low precision annotators where one is unbiased and the other is reasonably biased. The coefficients of determination ($R^2$) of the simulated annotators are: [0.940, 0.785, -2.469, -0.131, -1.749]. Computing the mean of the answers of the different annotators yields a $R^2$ of 0.798. Table~\ref{table:data_stats_reg} gives an overview on the statistics of datasets used in the regression experiments. 

\begin{table*}[t!]
\caption{Overall statistics of the regression datasets used in the experiments.}
\label{table:data_stats_reg}
\begin{center}
\begin{tabular}{c|c|c|c|c|c|c}
Dataset& \specialcell{Train/test\\sizes} & \specialcell{Annotators\\source}  & \specialcell{Num. answers per\\ instance ($\pm$ stddev.)} & \specialcell{Num. answers per\\ worker ($\pm$ stddev.)} & \specialcell{Mean annotators\\ $R^2$ ($\pm$ stddev.)}  & \specialcell{Mean answer\\ $R^2$}  \\
\hline
we8there & 4624/1542 & Simulated & 5.000 $\pm$ 0.000 & 4624.000 $\pm$ 0.000 & -0.525 $\pm$ 1.364 & 0.798 \\
movie reviews & 1500/3506 & Mech. Turk & 4.960 $\pm$ 0.196 & 55.111 $\pm$ 171.092 & -0.387 $\pm$ 1.267 & 0.830 \\
\end{tabular}
\end{center}
\vspace*{-0.3cm}
\end{table*}%

We compare the proposed model (MA-sLDAr) with the two following baselines:

\begin{itemize}[itemsep=0.02cm]
\item \textit{LDA + LinReg (mean)}: This baseline corresponds to applying unsupervised LDA to the data, and learning a linear regression model on the inferred topics distributions of the documents. The answers from the different annotators were aggregated by computing the mean.
\item \textit{sLDA (mean)}: This corresponds to using the regression version of sLDA \cite{Mcauliffe2008} with the target variables obtained by computing the mean of the annotators' answers.
\end{itemize}

\begin{figure}[!t]
\includegraphics[trim=0 1.1cm 0 0.5cm, clip, scale=.34]{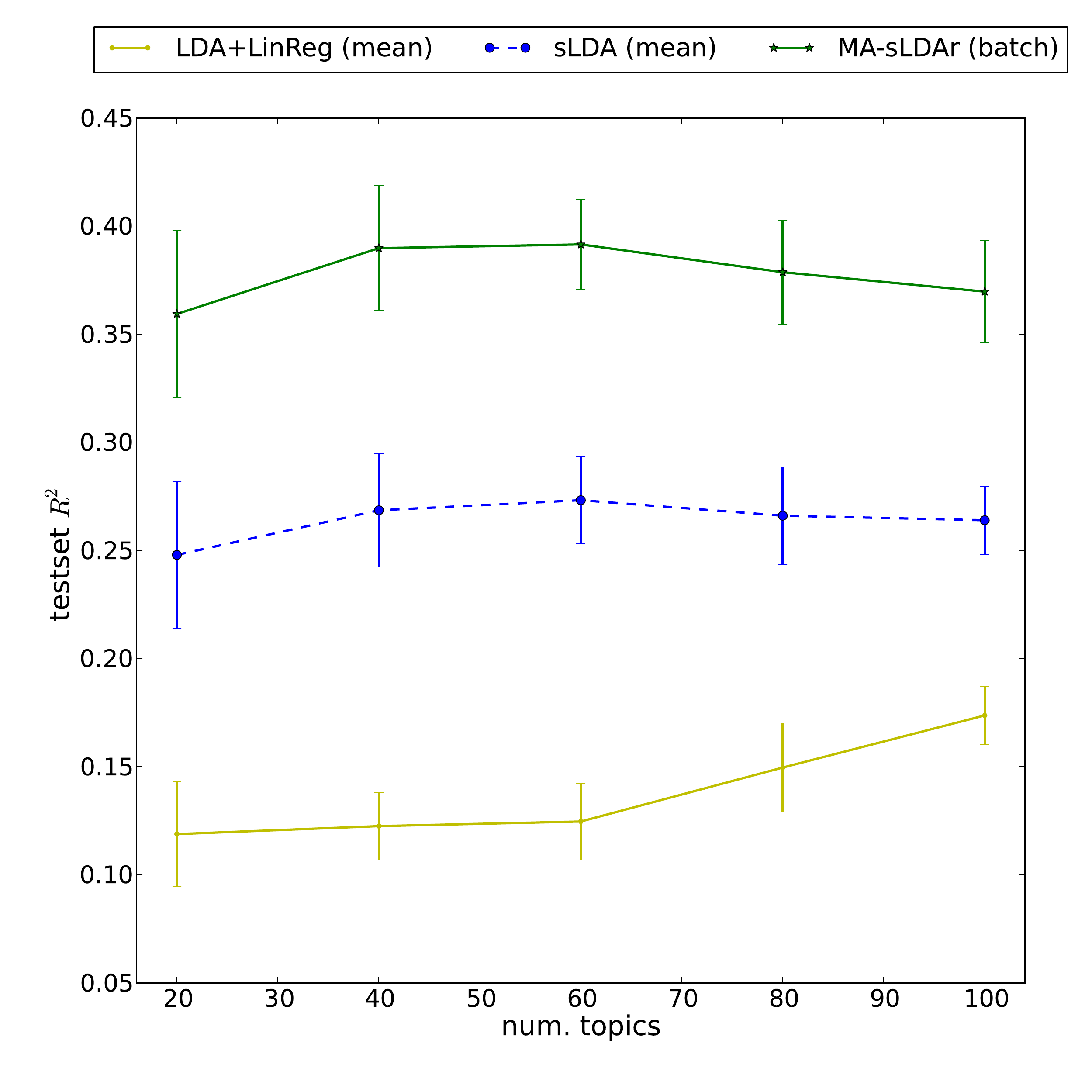}
\caption{Average testset $R^2$ (over 30 runs; $\pm$ stddev.) of the different approaches on the we8there data.}
\label{fig:we8there}
\end{figure}

Fig.~\ref{fig:we8there} shows the results obtained for different numbers of topics. Do to the stochastic nature of both the annotators simulation procedure and the initialization of the variational Bayesian EM algorithm, we repeated each experiment 30 times and report the average $R^2$ obtained with the corresponding standard deviation. Since the regression datasets that are considered in this article are not large enough to justify the use of a stochastic variational inference (svi) algorithm, we only made experiments using the batch algorithm developed in Section~\ref{subsec:approx_inf_reg}. The results obtained clearly show the improved performance of MA-sLDAr over the other methods. 

\subsubsection{Amazon Mechanical Turk}

\begin{figure}[t]
\centering
\includegraphics[width=8.5cm]{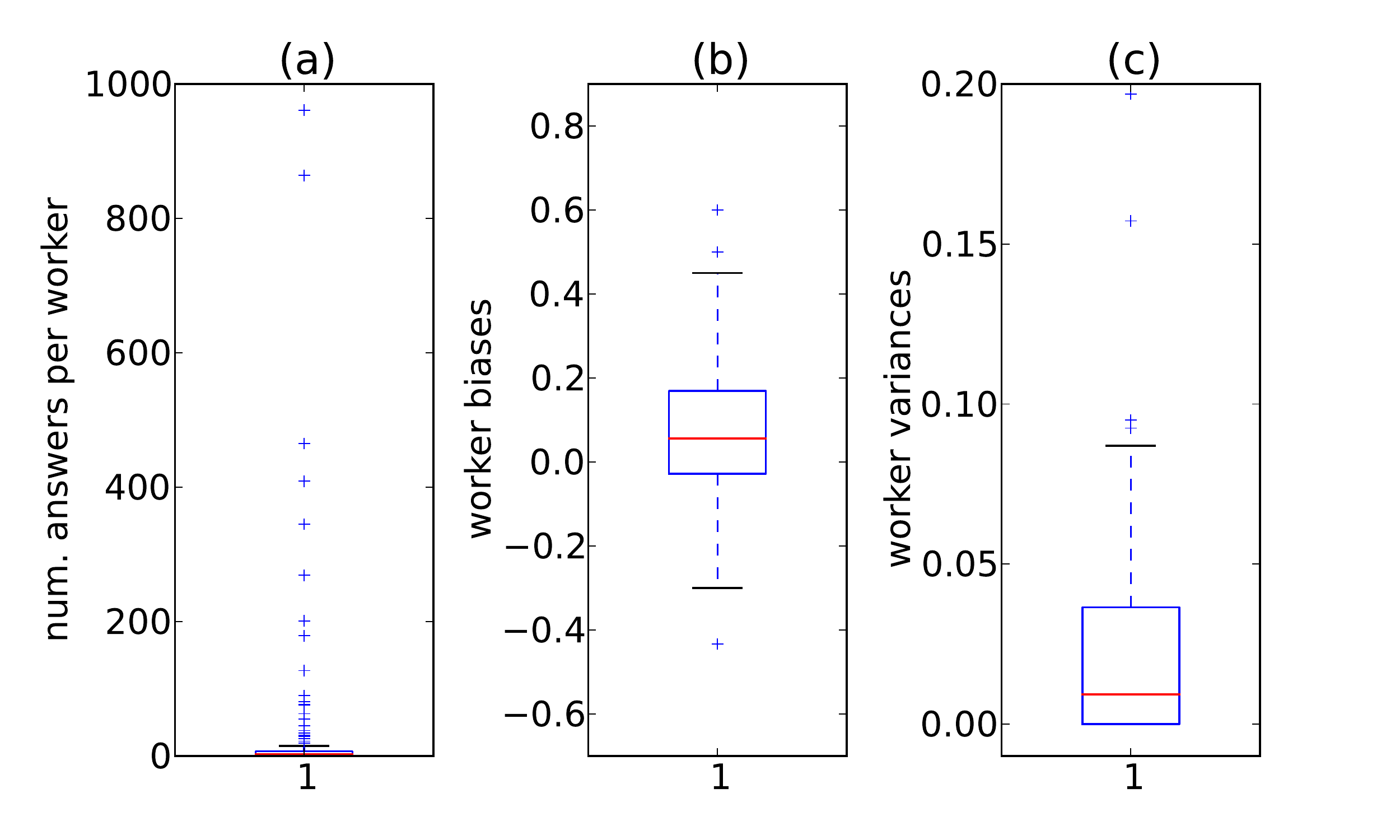}
\caption{Boxplot of the number of answers per worker (a) and their respective biases (b) and variances (c) for the movie reviews dataset.}
\label{fig:boxplot_moviereviews}
\end{figure}

The proposed multi-annotator regression model (MA-sLDAr) was also validated with real annotators by using AMT. For that purpose, the movie review dataset from \cite{Pang2005} was used. This dataset consists of 5006 movie reviews along with their respective star rating (from 1 to 10). The goal of this experiment is then predict how much a person liked a movie based on what she says about it. We ask workers to guess how much they think the writer of the review liked the movie based on her comments. An average of 4.96 answers per-review was collected for a total of 1500 reviews. The remaining reviews were used for testing. In average, each worker rated approximately 55 reviews. Using the mean answer as an estimate of the true rating of the movie yields a $R^2$ of 0.830. Table~\ref{table:data_stats_reg} gives an overview of the statistics of this data. Fig.~\ref{fig:boxplot_moviereviews} shows boxplots of the number of answers per worker, as well as boxplots of their respective biases ($b^r$) and variances (inverse precisions, $1/p^r$).

\begin{figure}[!t]
\includegraphics[trim=0 1.1cm 0 0.5cm, clip, scale=.34]{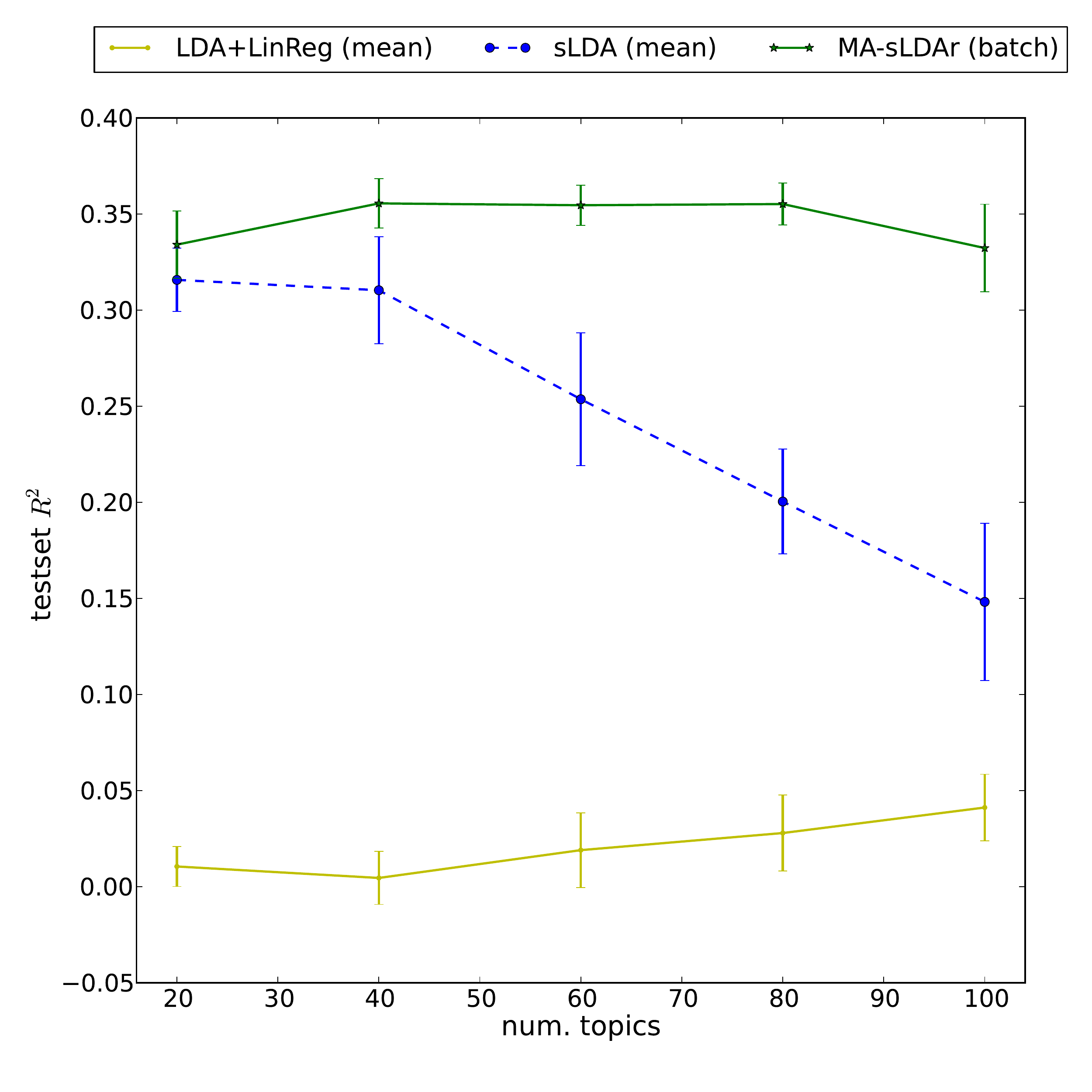}
\caption{Average testset $R^2$ (over 30 runs; $\pm$ stddev.) of the different approaches on the movie reviews data.}
\label{fig:moviereviews}
\vspace*{-0.3cm}
\end{figure}

The preprocessing of the text consisted of stemming and stop-words removal. Using the preprocessed data, the proposed MA-sLDAr model was compared with the same baselines that were used with the we8there dataset in Section~\ref{subsec:we8there}. Fig.~\ref{fig:moviereviews} shows the results obtained for different numbers of topics. These results show that the proposed model outperforms all the other baselines. 

\begin{figure}[!t]
\centering
\includegraphics[trim= 0 2.4cm 0 7.2cm, clip, scale=.25]{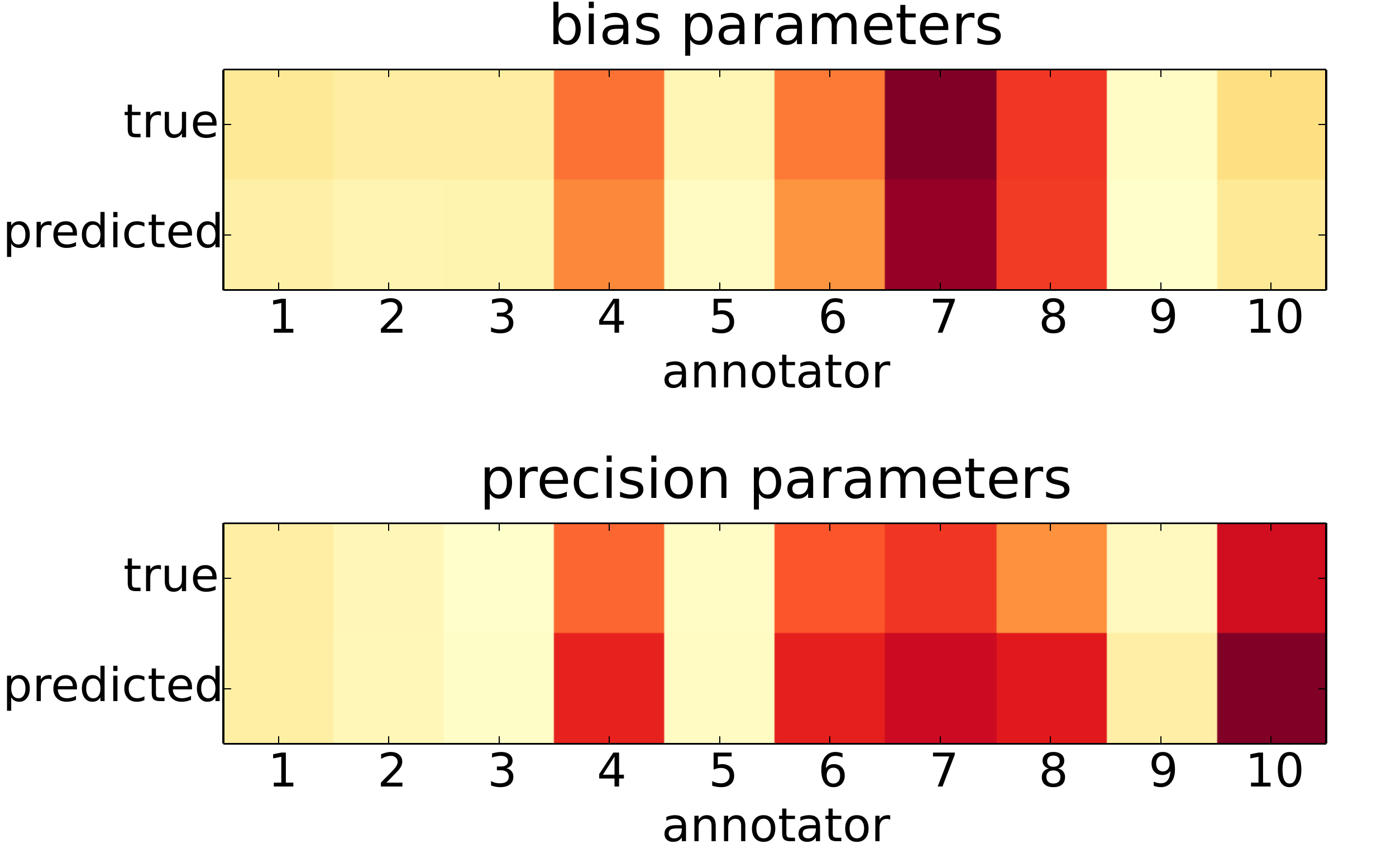}
\caption{True vs. predicted biases and precisions of 10 random workers of the movie reviews dataset.}
\label{fig:plot_b_v}
\end{figure}

With the purpose of verifying that the proposed model is indeed estimating the biases and precisions of the different workers correctly, we plotted the true values against the estimates of MA-sLDAr with 60 topics for a random subset of 10 workers. Fig.~\ref{fig:plot_b_v} shows the obtained results, where higher color intensities indicate higher values. Ideally, the colour of two horizontally-adjacent squares would then be of similar shades, and this is indeed what happens in practice for the majority of the workers, as Fig.~\ref{fig:plot_b_v} shows. Interestingly, the figure also shows that there are a couple of workers that are considerably biased (e.g. workers 6 and 8) and that those biases are being correctly estimated, thus justifying the inclusion of a bias parameter in the proposed model, which contrasts with previous works \cite{Raykar2010,Groot2011}.

\section{Conclusion}
\label{sec:conclusion}

This article proposed a supervised topic model that is able to learn from multiple annotators and crowds, by accounting for their biases and different levels of expertise. Given the large sizes of modern datasets, and considering that the majority of the tasks for which crowdsourcing and multiple annotators are desirable candidates, generally involve complex high-dimensional data such as text and images, the proposed model constitutes a strong contribution for the multi-annotator paradigm. This model is then capable of jointly modeling the words in documents as arising from a mixture of topics, as well as the latent true target variables and the (noisy) answers of the multiple annotators. We developed two distinct models, one for classification and another for regression, which share similar intuitions but that inevitably differ due to the nature of the target variables. We empirically showed, using both simulated and real annotators from Amazon Mechanical Turk that the proposed model is able to outperform state-of-the-art approaches in several real-world problems, such as classifying posts, news stories and images, or predicting the number of stars of restaurant and the rating of movie based on their reviews. For this, we use various popular datasets from the state-of-the-art, that are commonly used for benchmarking machine learning algorithms. Finally, an efficient stochastic variational inference algorithm was described, which gives the proposed models the ability to scale to large datasets. 


\ifCLASSOPTIONcompsoc
  \section*{Acknowledgments}
\else
  \section*{Acknowledgment}
\fi

{The Funda\c{c}\~ao para a Ci\^encia e Tecnologia (FCT) is gratefully acknowledged for founding this work with the grants SFRH/BD/78396/2011 and PTDC/ECM-TRA/1898/2012 (InfoCROWDS).}

\ifCLASSOPTIONcaptionsoff
  \newpage
\fi



\bibliographystyle{IEEEtran}
\bibliography{ma-slda_v2}
%
%
%

%

\begin{IEEEbiography}[{\includegraphics[width=1in,height=1.25in,clip,keepaspectratio]{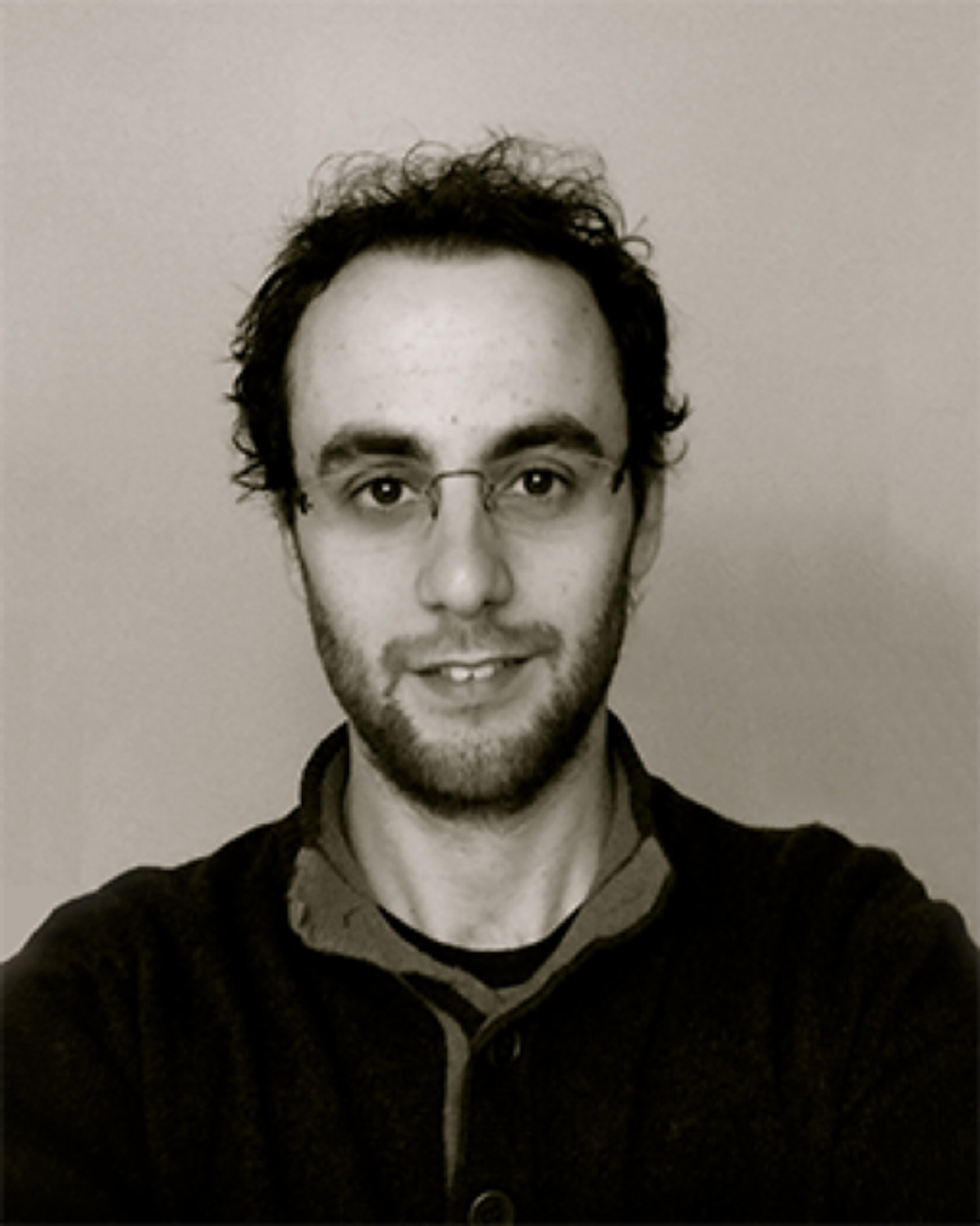}}]{Filipe~Rodrigues}
is Postdoctoral Fellow at Technical University of Denmark, where he is working on machine learning models for understanding urban mobility and the behavior of crowds, with emphasis on the effect of special events in mobility and transportation systems. He received a Ph.D. degree in Information Science and Technology from University of Coimbra, Portugal, where he developed probabilistic models for learning from crowdsourced and noisy data. His research interests include machine learning, probabilistic graphical models, natural language processing, intelligent transportation systems and urban mobility. 
\end{IEEEbiography}

\begin{IEEEbiography}[{\includegraphics[width=1in,height=1.25in,clip,keepaspectratio]{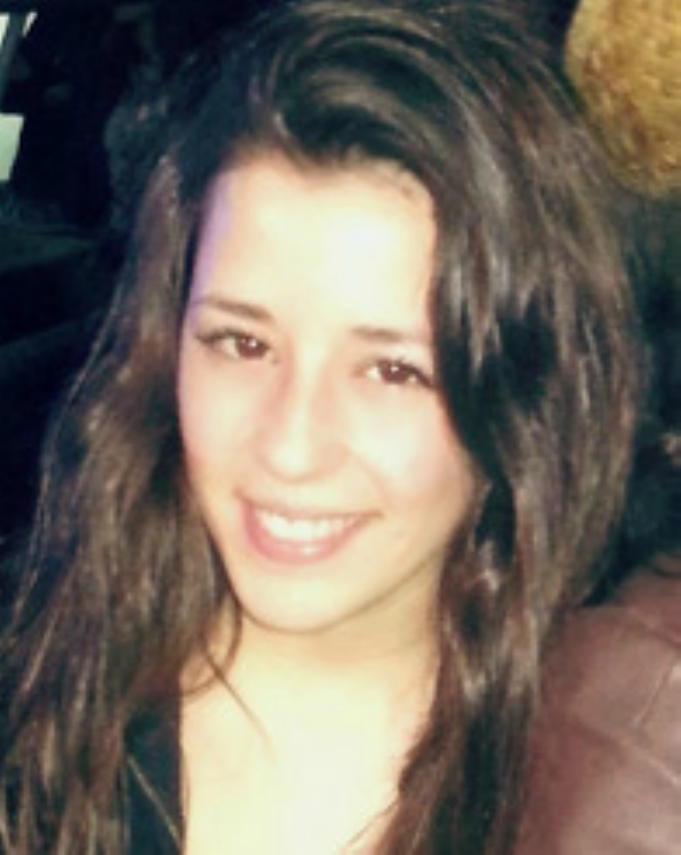}}]{Mariana~Louren\c{c}o}
has a MSc degree in Informatics Engineering from University of Coimbra, Portugal. Her thesis presented a supervised topic model that is able to learn from crowds and she took part in a research project whose primary objective was to exploit online information about public events to build predictive models of flows of people in the city. Her main research interests are machine learning, pattern recognition and natural language processing. 
\end{IEEEbiography}

\begin{IEEEbiography}[{\includegraphics[width=1in,height=1.25in,clip,keepaspectratio]{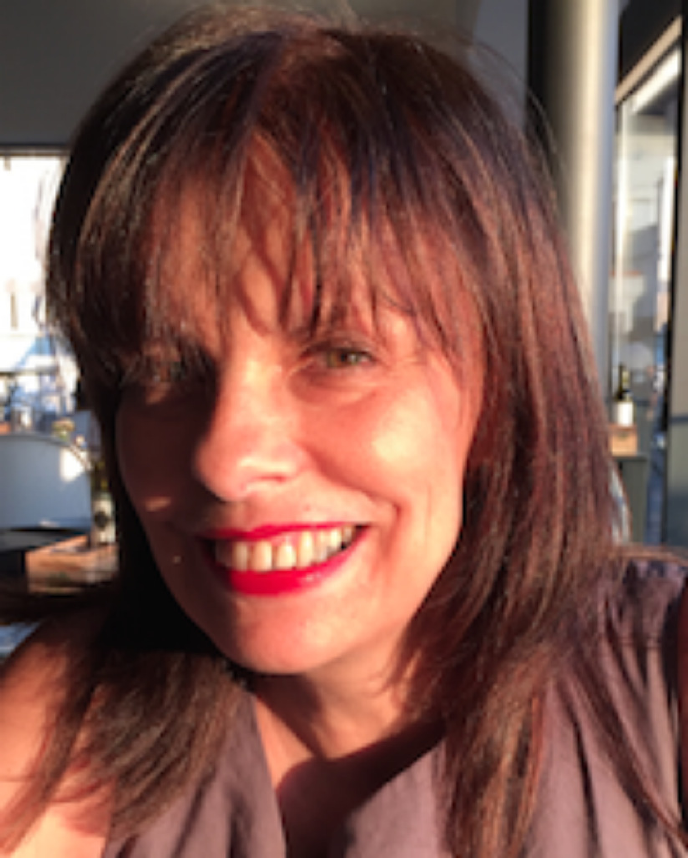}}]{Bernardete~Ribeiro}
is Associate Professor at the Informatics Engineering Department, University of Coimbra in Portugal, from where she received a D.Sc. in Informatics Engineering, a Ph.D. in Electrical Engineering, speciality of Informatics, and a MSc in Computer Science. Her research interests are in the areas of Machine Learning, Pattern Recognition and Signal Processing and their applications to a broad range of fields. She was responsible/participated in several research projects in a wide range of application areas such as Text Classification, Financial, Biomedical and Bioinformatics. Bernardete Ribeiro is IEEE Senior Member, and member of IARP International Association of Pattern Recognition and ACM.
\end{IEEEbiography}

\begin{IEEEbiography}[{\includegraphics[width=1in,height=1.25in,clip,keepaspectratio]{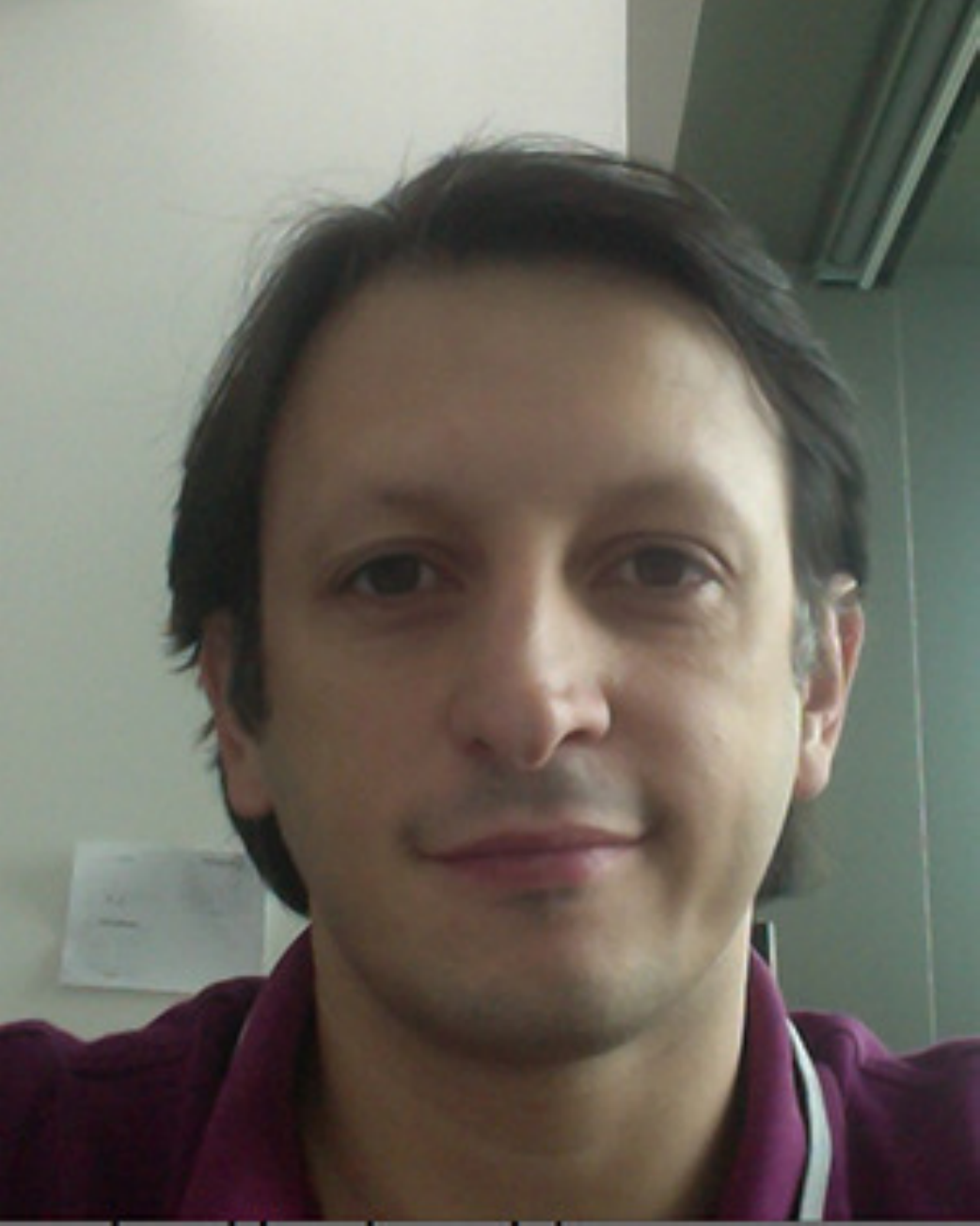}}]{Francisco~C.~Pereira}
is Full Professor at the Technical University of Denmark (DTU), where he leads the Smart Mobility research group. His main research focus is on applying machine learning and pattern recognition to the context of transportation systems with the purpose of understanding and predicting mobility behavior, and modeling and optimizing the transportation system as a whole. He has Master€™s (2000) and Ph.D. (2005) degrees in Computer Science from University of Coimbra, and has authored/co-authored over 70 journal and conference papers in areas such as pattern recognition, transportation, knowledge based systems and cognitive science. Francisco was previously Research Scientist at MIT and Assistant Professor in University of Coimbra. He was awarded several prestigious prizes, including an IEEE Achievements award, in 2009, the Singapore GYSS Challenge in 2013, and the Pyke Johnson award from Transportation Research Board, in 2015.
\end{IEEEbiography}




\end{document}